\documentclass[letterpaper, 10 pt, conference]{ieeeconf}    % use this command before submit
\IEEEoverridecommandlockouts                                % This command is only needed if
\usepackage{multirow}
\usepackage{algorithm}
\usepackage{cite}
\usepackage{algorithmic}
\usepackage{graphicx}
\usepackage{times}
\usepackage{amsmath}
\usepackage{amssymb}
\usepackage{subfigure}
\usepackage{graphics} % for pdf, bitmapped graphics files
\usepackage[utf8]{inputenc}
\usepackage{boldline}
\usepackage{multicol}
\usepackage{lipsum}
\usepackage{booktabs}

\usepackage[pdfstartview=FitH,bookmarksnumbered=true,colorlinks,bookmarksopen=true]{hyperref}
\newcommand{\etal}{\textit{et al}.}

\newcommand{\eg}{\textit{e}.\textit{g}.}
\title{\LARGE \bf
Robust Robotic Pouring using Audition and Haptics}
\author{Hongzhuo~Liang$^{1*}$, Chuangchuang~Zhou$^{1,2}$, Shuang~Li$^{1}$, Xiaojian~Ma$^{3}$, \\Norman~Hendrich$^{1}$, Timo~Gerkmann$^{4}$, Fuchun~Sun$^{5}$, Marcus Stoffel$^{2}$, and Jianwei Zhang$^{1}$% <-this % stops a space
\thanks{This research was funded by the German Research Foundation (DFG)
and the National Science Foundation of China (NSFC)
in project Crossmodal Learning, DFG TRR-169/NSFC 61621136008,
and partially supported by European projects H2020 STEP2DYNA (691154) and Ultracept (778602).}
\thanks{$^{*}$Corresponding author, email: liang@informatik.uni-hamburg.de}%
\thanks{$^{1}$Group TAMS, Dept. of Informatics, Universit\"{a}t Hamburg.}% <-this % stops a space
\thanks{$^{2}$Institute of General Mechanics, RWTH Aachen University.}% <-this % stops a space
\thanks{$^{3}$Group VCLA, Dept. of Statistics, University of California, Los Angeles.}% <-this % stops a space
\thanks{$^{4}$Signal Processing (SP), Dept. of Informatics, Universit\"{a}t Hamburg.}% <-this % stops a space
\thanks{$^{5}$Dept. of Computer Science and Technology, Tsinghua University.}% <-this % stops a space
}
\begin{document}
\maketitle
\thispagestyle{empty}
\pagestyle{empty}

%%%%%%%%%%%%%%%%%%%%%%%%%%%%%%%%%%%%%%%%%%%%%%%%%%%%%%%%%%%%%%%%%%%%%%%%%%%%%%%%
\begin{abstract}
Robust and accurate estimation of liquid height lies as an essential part of pouring tasks for service robots.
However, vision-based methods often fail in occluded conditions while audio-based methods cannot work well in a noisy environment.
We instead propose a multimodal pouring network (MP-Net) that is able to robustly predict liquid height by conditioning on both audition and haptics input.
MP-Net is trained on a self-collected multimodal pouring dataset.
This dataset contains 300 robot pouring recordings with audio and force/torque measurements for three types of target containers.
We also augment the audio data by inserting robot noise.
We evaluated MP-Net on our collected dataset and a wide variety of robot experiments. 
Both network training results and robot experiments demonstrate that MP-Net is robust against noise and changes to the task and environment.
Moreover, we further combine the predicted height and force data to estimate the shape of the target container.
\end{abstract}

\section{Introduction}
Pouring a specific amount of liquid into a container is an important manipulation skill for service robots in applications such as bartending or housekeeping and also in industrial environments.
Compared to humans who can effortlessly pour the liquid into a container without spilling,
predicting the liquid height in the target container is still challenging for a robot.
For the robust and accurate perception of robot pouring, recent approaches mainly rely on vision~\cite{visualpouring2, visualpouring3, visualpouring4}. They take the RGB image as input, then infer the liquid height, liquid amount, even pouring trajectories.
However, vision fails under occlusions or in the dark.
Using audio~\cite{audio_corl, wilson2019analyzing, liang2019AudioPouring} is a second option for robot pouring perception,
but the performance of this method will degrade in a noisy environment.
Besides, if the robot is equipped with force/torque sensing, the amount of liquid poured out from the source container can be measured directly~\cite{huang2017learning}, 
but the liquid height in a target container can only be predicted if the initial fill level and the shape of the container are known.
These drawbacks in existing perception methods suggest that
to estimate the liquid height precisely, it is necessary
to find a robust approach that still works in conditions where one modality
becomes incomplete or deteriorated.

Therefore, we propose to tackle the issues of robust robotic pouring by a multimodal perception method.
This multimodal system should be robust in a wild environment, for example, against different levels of noise, types of noise, or light conditions.
The perception system should be able to generalize to different target containers, liquid types, and initial liquid height in the target containers.

\begin{figure}[t]
    \centering
    \includegraphics[width=0.486\textwidth]{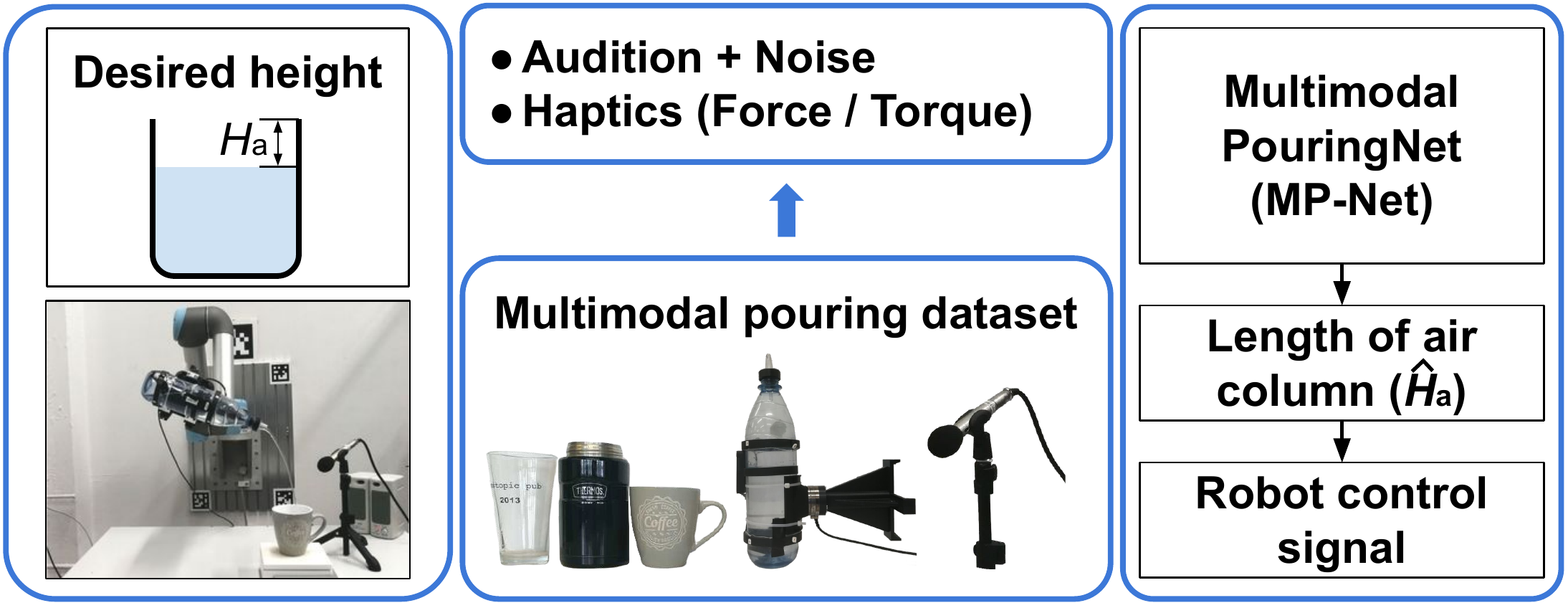}
    \caption{Our multimodal pouring pipeline. (Left) The goal of our pouring task is to pour the desired amount of liquid into a target container.
The generated sounds depend on the height of the remaining air column ($H_a$) in the container.
(Center) We collected a multimodal robotic pouring dataset, which contains 300 pouring sequences with audio and force/torque sensor data for training our network. A scale recorded the weight of the poured liquid. We then calculated $H_a$ to serve as ground truth. (Right) By giving the audio and haptic as input, our Multimodal PouringNet (MP-Net) predicts $\hat{H}_a$ as output. This output is then used to control the robotic pouring.}
    \vskip -0.15in
    \label{fig:mp_pipeline}
\end{figure}

In this paper, we develop a novel multimodal neural network called MP-Net that takes preprocessed audio spectrograms and force/torque data as the input and estimates the length of the air column in the target container.
This idea is inspired by human 
experiences, as 
humans can infer whether the target container is almost full from the pouring sound 
and use their proprioceptive haptic feedback to estimate how much liquid is poured out from the source container~\cite{ikeno2015change, pouring-dark}.
Fig.~\ref{fig:mp_pipeline} illustrates the proposed perception pipeline.

The main structure of MP-Net is a recurrent neural network that implicitly integrates the prior knowledge that the liquid level rises monotonically during the pouring.
A multimodal pouring dataset with audio and haptics was built up to train this model. 
To further improve the adaptability to the noisy environment of MP-Net, we augmented the audio data by adding different noise levels of robot noise. 
The benefits of force/torque data and audio data augmentation are verified in both our dataset and a wide range of robotic experiments, respectively.

To sum up, our main contributions are:
\begin{enumerate}
    \item We propose a multimodal neural network MP-Net to estimate the liquid height robustly and to enhance the precision and generalization ability of robotic pouring.
    Not only the haptic input but also the audio data augmentation facilitate to produce a robust model that can generalize broadly.
    \item System assessment and comparison across eight robotic experiments, \eg, pouring water into different containers, in various noisy environments with different noise levels and noise sources, revealed that high accuracy could be achieved by our proposed method.
    \item The shape of the target container is generated by combining the force data and realtime height prediction.
\end{enumerate}

\section{Related Work}
\textbf{Visual sensing for robotic pouring.}
Vision remains the most popular modality in robot pouring perception,
as humans also highly rely on vision to perform the same task.
Pithadiya \etal~\cite{pithadiya2011selecting} compared several edge detection techniques for the filling height inspection of convex target containers.
Schenck \etal~\cite{visualpouring1} used a thermal camera to generate pixel-level ground truth data of heated water.
The RGB camera is then used to determine the water volume using both a model-based method and a neural network method.
This method suffers from poor estimation accuracy due to the varied liquid types and the complex shapes of the water flow.
To predict the absolute height of the liquid, Dong \etal~\cite{Dong2019DepthPouring} used a point cloud to model the target container and a proportional-derivative controller to perform the pouring action.
Do \etal~\cite{dochau2019} utilized a Kalman filter dealing with opaque and transparent liquids based on an RGB-D camera.
However, the mean height errors for ten pourings of three transparent liquids were larger than 4\,mm.
Recently, \cite{dochau2018learning, sermanet2016unsupervised}  deployed reinforcement learning methods for generating motion trajectories during the pouring process.

\textbf{Audio sensing for robotic pouring.}
The auditory signal generated when liquid or granular materials interact with other objects
also comprises reliable clues, such as resonance frequency and vibration behavior of the liquids and the air.
Griffith \etal~\cite{griffith2012object} indicated that auditory and proprioceptive data enhance the classification tasks for the interaction between objects and water.
Clarke \etal~\cite{audio_corl} used audio-frequency vibration generated by shaking a granular material to evaluate the weight that is poured out.
However, the weight of the poured out granular materials does not allow us to estimate the filling height of the target container.

None of the above work explored the fact that the audio vibration of the air in the target container
can be analyzed to solve the liquid height regression in robotic pouring.
In our previous work~\cite{liang2019AudioPouring}, we used audio as input and trained an RNN based network to directly predict the length of the air column, including generalization to different target containers.
However, the network failed in a noisy environment or when pouring liquids with high viscosity
as the audio signal is weak in these conditions.

\textbf{Haptic sensing for robotic pouring.}
Haptic sensing, especially force and torque sensing, is also an essential signal of robotic pouring.
The goal of learning robust robot pouring from human demonstrations including hand trajectories and force/torque data is explicitly discussed by Huang and Sun ~\cite{huang2019dataset}, as part of their multi-task dataset on human daily interactive manipulation.
Rozo \etal~\cite{rozo2013force} used a parametric hidden Markov model to retrieve joint-level commands given the force/torque inputs from the human demonstration.
Saal \etal~\cite{saal2010active} examined the viscosity estimation of various liquids inside a bottle from tactile sensory data.
Matl \etal~\cite{matl2019haptic} utilized a force sensor mounted at the end of a robotic arm with a container grasped by a gripper to get the change of
wrenches, and they used a physics-based model to estimate the mass and volume of the liquid.
In the above three papers, perception happens before the pouring action begins.
Although the force signal from the pouring container can explicitly represent the volume of the poured-out liquid,
it cannot measure the liquid height in an unseen target container.

\begin{figure}[t]
    \centering
    \includegraphics[width=0.486\textwidth]{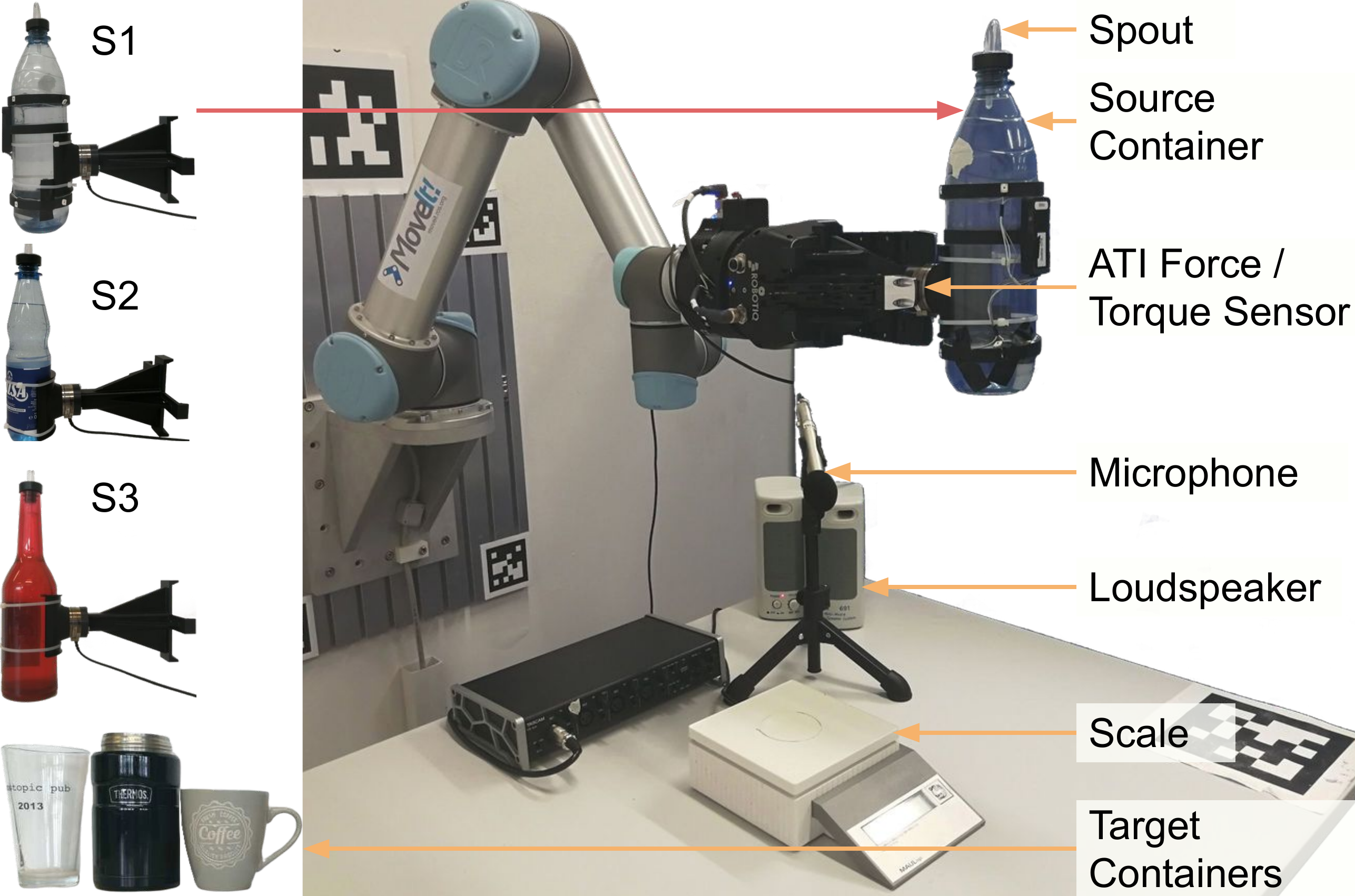}
    \caption{Pouring setup used to collect our multimodal dataset and to perform the robot experiment.
The three target containers used to collect the dataset are shown in the bottom left corner.
We put these containers on a calibrated scale to get the realtime label of the liquid height. A force/torque sensor is mounted on the source container to collect force and torque information while pouring, and a microphone was set on the table to collect the pouring audio.
Loudspeakers were used to create ambient noise for robot experiments.
S1, S2, S3 in the left represent three different source containers where S1 is the container used to collect dataset, S2, and S3 are novel source containers.
}
    \vskip -0.25in
    \label{fig:dataset_setup}
\end{figure}

\textbf{Multimodal fusion for robotic pouring.}
Recent studies have shown that multimodal sources represent the environmental features better than single modality in pouring tasks \cite{sanchez2020} as weel as other tasks~\cite{gan2019look}.
Wu \etal~\cite{wu2018liquid} presented a hierarchical long short-term memory~\cite{hochreiter1997long} model, which could detect if a pouring sequence was successful or failed. This model is based on a pouring dataset, including visual sequences and IMU data.
However, this work only classifies whether a pouring action was successful but cannot predict the precise height of the poured-out liquid.
Wilson \etal~\cite{wilson2019analyzing} implemented a multimodal convolutional neural network to fuse audio and visual data to predict the weight of the poured liquid, detect overflow and to classify the liquid and the target container.
Park \etal~\cite{multimodal-anomaly-detection} proposed another interesting robot application for anomaly detection during robot manipulation using haptic, visual, auditory, and kinematic sensing.

Based on how humans rely on the correlations of haptic and audio while pouring, in this paper, we take advantage of audio and force/torque data as the input to promote the robustness of robotic pouring.

\section{Multimodal Pouring Dataset}
\subsection{Dataset Collection Setup}
To create the multimodal dataset for the height estimation task, we designed a robot pouring setup, as shown in Fig.~\ref{fig:dataset_setup}.
It consists of a UR5 robot arm with a Robotiq 3-finger hand,
and a custom 3D-printed bottle holder with an embedded ATI force/torque sensor.
A standard plastic bottle (S1) is used as the source container, with an optional bottle spout to limit liquid flow.
The target container is placed on a MAULlogic digital scale which measures the combined weight of the container and the liquid inside.
Audio is recorded with a microphone in an environment with only UR5 robot ego-noise.
The dataset setup contains three different target containers with three different materials and heights.
The properties of these containers are shown in Table~\ref{tab:containers} with ID 1-3.

Unlike our previous work that used human pouring sequences as training data~\cite{liang2019AudioPouring},
we use the robot to perform the pouring task.
The main reason is that the haptic data coming from human pouring is massively different from the robot pouring data, which makes it hard to transform human pouring to robotic pouring.

We set a fixed pouring trajectory and place the mouth center of the source container 310\,mm above the scale plane.
In this manner, we collected 100 trials each for three different target containers.
For each trial, we recorded both audio and force/torque data starting just before the scale reading begins to change, and stopping after the reading of the scale becomes stable again.
The lengths of one pouring recording varied from 24-40 seconds according to
the capacity of the different target containers.

\subsection{Data Analysis}

\textbf{Audio-frequency data.}
To better extract audio information from the microphone,
we resampled all audio data from 44.1\,kHz to 16\,kHz and computed spectrograms with a window length of 32\,ms and 50\% overlap without zero-padding.
For network training, we randomly chose 4 seconds of audio clips from one complete pouring audio sequence.
Using fixed length training samples allows for batch processing for better GPU performance,
and random starts correspond to different initial liquid amounts,
so that the network can generalize to partially filled containers.

\begin{figure}[t]
    \centering
    \includegraphics[width=0.486\textwidth]{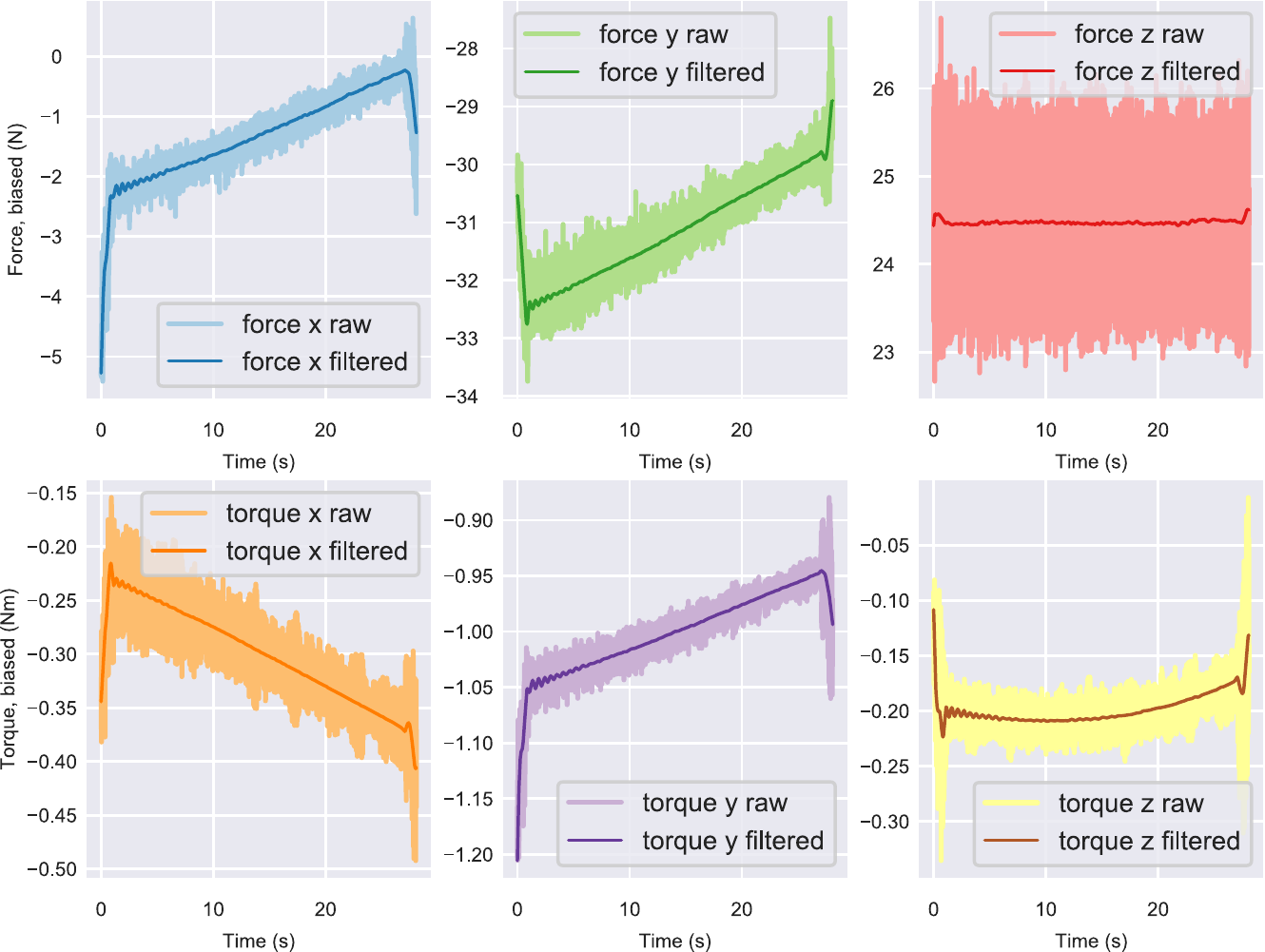}
    \vskip -3mm
    \caption{A sample of force/torque data (N/Nm) in our pouring dataset. In each sub-figure, we show the raw sensor data (light line) and the filtered data using a Butterworth low pass filter (dark line). See the text for details.}
    \vskip -0.5mm
    \label{fig:ft_vis_data}
\end{figure}

%\noindent
\textbf{Force/Torque data.}
The force/torque sensor we use is ATI Mini45 (500\,Hz).
As the weight of the source container is far below the maximum payload of this sensor (580\,N in Fx/Fy),
the raw data is rather noisy but becomes usable after simple low-pass filtering.
See Fig.~\ref{fig:ft_vis_data} for typical raw force/torque data (in N/Nm) during a sample pouring sequence.
In each sub-figure, we plot the noisy raw signal from the sensor shown as a light color
together with a low-pass filtered signal shown as a dark color.
The change of the bottle weight during pouring can clearly be seen in the force/torque readings
for the x- and y-axis, while the z-axis measurements show only little change.
This is due to the specific orientation of the sensor in our 3D-printed holder,
dictated by the cable routing.
While the sensor z-axis is almost horizontal (orthogonal to gravity),
the x- and y-axes of the sensor are pointing diagonally upwards,
so that the raw sensor readings actually increase when pouring from the source container.
The pouring motion from the robot slightly rotates the bottle around the z-axis.

\begin{figure}[t]
    \centering
    % \subfigure[]
    {\includegraphics[height=0.18\textwidth]{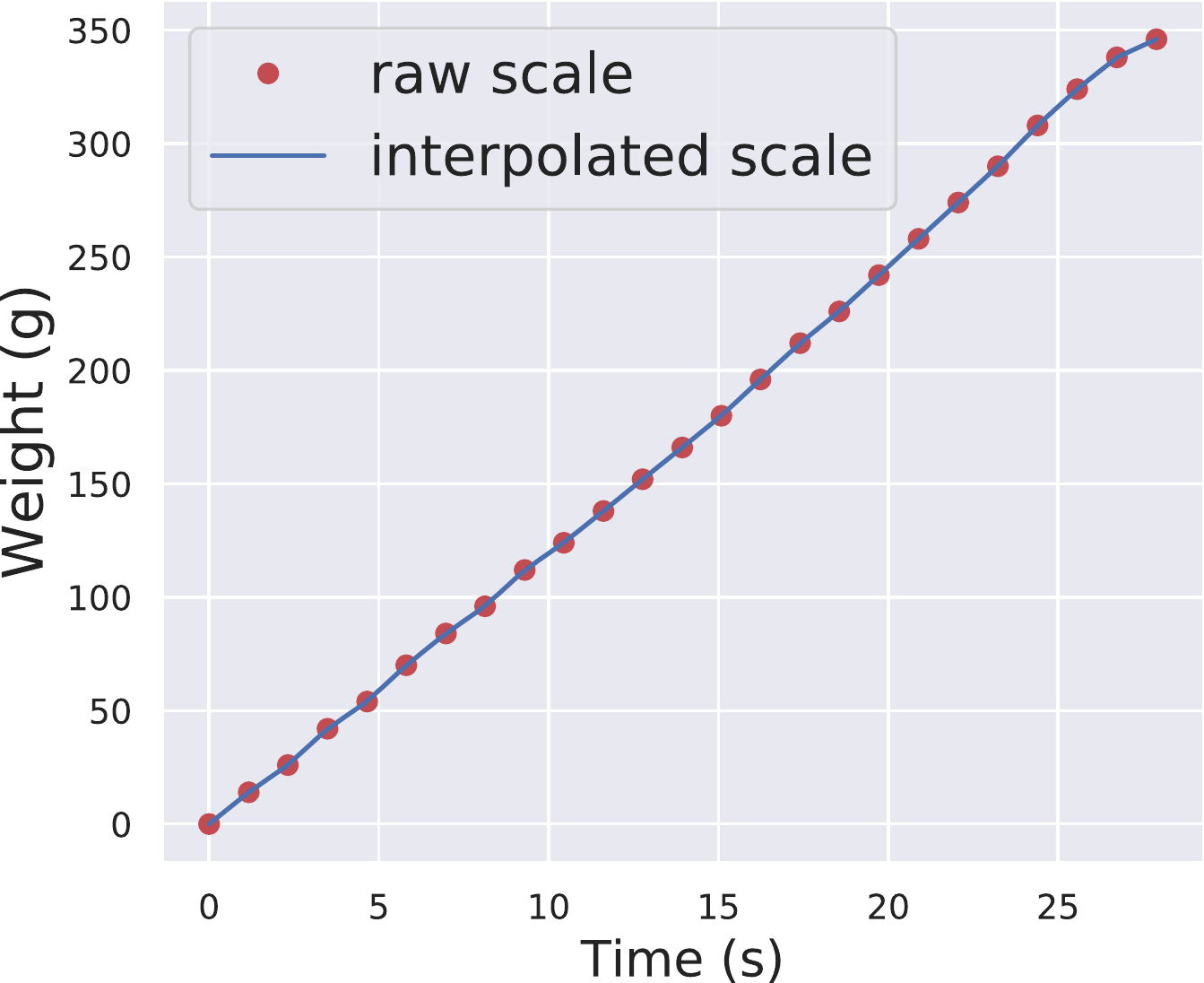}
    % \label{fig:scale_reading}
    }
     \hspace{2mm}
    % \subfigure[]
    {\includegraphics[height=0.18\textwidth]{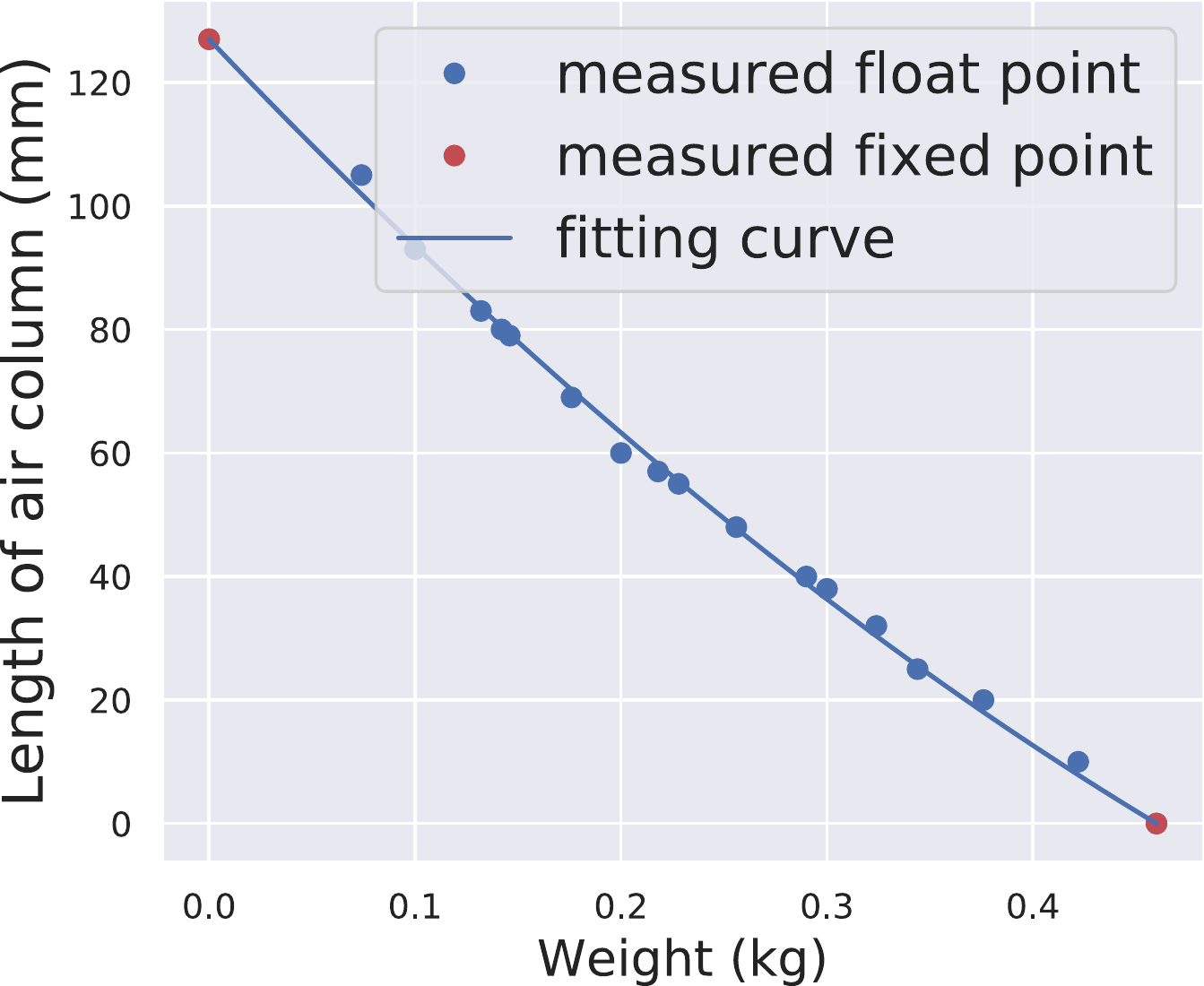}
    % \label{fig:weight2height}
    }
    \vskip -1mm
    \caption{%\subref{fig:scale_reading} 
    (a, left) Sample scale readings of a pouring sequence in our dataset from bottle ID 1. Each red dot is one weight measurement from the scale. The blue line is the interpolated curve.
    %\subref{fig:weight2height}
    (b, right) Converting liquid weights (kg) into length of the air column (mm) for bottle ID 1. The dotted points are the manually measured data, where the red points indicate the empty and completely filled container. The blue line is the fitting curve that calculates from the scale reading to the length of the air column.}
    \label{fig:scale_data_process}
    \vskip -4mm
\end{figure}

\begin{figure*}[t]
    \centering
    \includegraphics[width=1\textwidth]{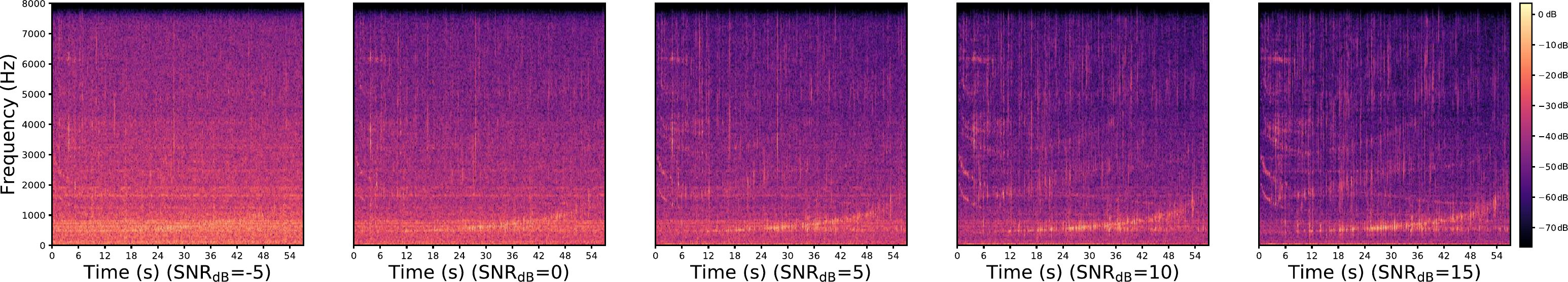}
    \caption{
    Examples of audio spectrograms that add different noise levels of the same audio signal in our dataset.
    While the $\rm SNR_{dB}$ is high (\eg, $\rm SNR_{dB}=15$, right), which indicates the audio signal is rather clear, we can clearly see rising frequency curves between 256\,Hz-2048\,Hz. But when the $\rm SNR_{dB}$ is low (\eg, $\rm SNR_{dB}=-5$, left), there is no meaningful structure of the resonance frequency.}
    \vskip -0.15in
    \label{fig:audio}
\end{figure*}

%\noindent
\textbf{Scale data.}
To get the ground truth of the pouring perception problem, we used a MAULlogic digital scale
to measure the weight of the target container (accuracy $\pm 2$\,g).
Because the publishing frequency of the scale is only 1\,Hz,
we deployed a linear interpolation (Fig.~\ref{fig:scale_data_process}(a))
to get realtime scale readings.
To convert the measured weights into the heights $H_a$ of the air column needed for network training,
we sampled 10-15 random amounts of liquid for each target container, up to full container capacity,
and manually measured the length of the air column.
A polynomial curve fitting is utilized using this data to calculate the approximate air column length for different amounts of liquid and corresponding measured weights (Fig.~\ref{fig:scale_data_process}(b)).

\subsection{Audio Data Augmentation}
Considering that the variability of the audio data is significant enough, models trained on it will better generalize in different audio conditions.
Therefore, we augmented the audio data in our dataset by adding noise.
We recorded an ego noise from a humanoid PR2 robot, which is commonly used for research in household environments.
%By adding noise into the dataset, the model could have the potential to generalize to a noisy environment.
In detail, we employed the signal-to-noise ratio in decibels (${\rm SNR_{dB}}$) as our reference on how much noise was added.
\begin{equation}
    \label{eq1}
    {\rm SNR_{dB}} = 10 \log\left( \frac{A_{\rm signal}}{ \alpha_{\rm nf} \times A_{\rm noise} }\right)^2
\end{equation}
\begin{equation}
    \label{eq2}
    \alpha_{\rm nf} = \frac{A_{\rm signal}}{A_{\rm noise}} \times 10^{\frac{{\rm SNR_{dB}}}{-20}}
\end{equation}
\begin{equation}
    \label{eq3}
    \rm audio = signal + noise \times \alpha_{nf}
\end{equation}
Equations~(\ref{eq1}) - (\ref{eq3}) illustrate how noise was added,
where $A_{\rm signal}$ and $A_{\rm noise}$ are the root mean square (RMS) amplitude for the recorded pouring audio and noise, respectively.
$\alpha_{\rm nf}$ is the ratio of the input noise.

We mixed the original audio sequences with noise
of dif\-fer\-ent levels ranging from -20-20 $\rm SNR_{dB}$ with a step size of 5\,dB.
The number of generated audio clips, of 4 seconds length each, is proportional to the length of a pouring trial.
Fig.~\ref{fig:audio} shows examples of mixed audio spectro\-grams for different $\rm SNR_{dB}$.
When $\rm SNR_{dB} > 0$, one high-energy and rising curve between 256\,Hz and 2048\,Hz is clearly visible during pouring.
This curve represents the resonance frequency of the air. 
But when $\rm SNR_{dB} \le 0$, the spectro\-gram has little structure of the resonance frequency, which indicates this audio sequence contains more noise than signal.

\section{Multimodal Pouring Network}\label{sec:net}
% check this part for reason to use lstm and the phiscal model.
Our goal is to design a robust network architecture to acquire the liquid filling height by making sense of audition and haptics together.
Audio vibration results in the changes in the resonance frequency of the air in the target container.
Force/torque changes result from the liquid poured out of the source container, which yields additional guidance about the liquid in the target container, especially when the audio signal is deteriorated.
% By fusing these two modalities, we believe that the network could predict liquid height by conditioning on both audio and/or haptic input.

\begin{figure}[t]
    \centering
    \includegraphics[width=0.486\textwidth]{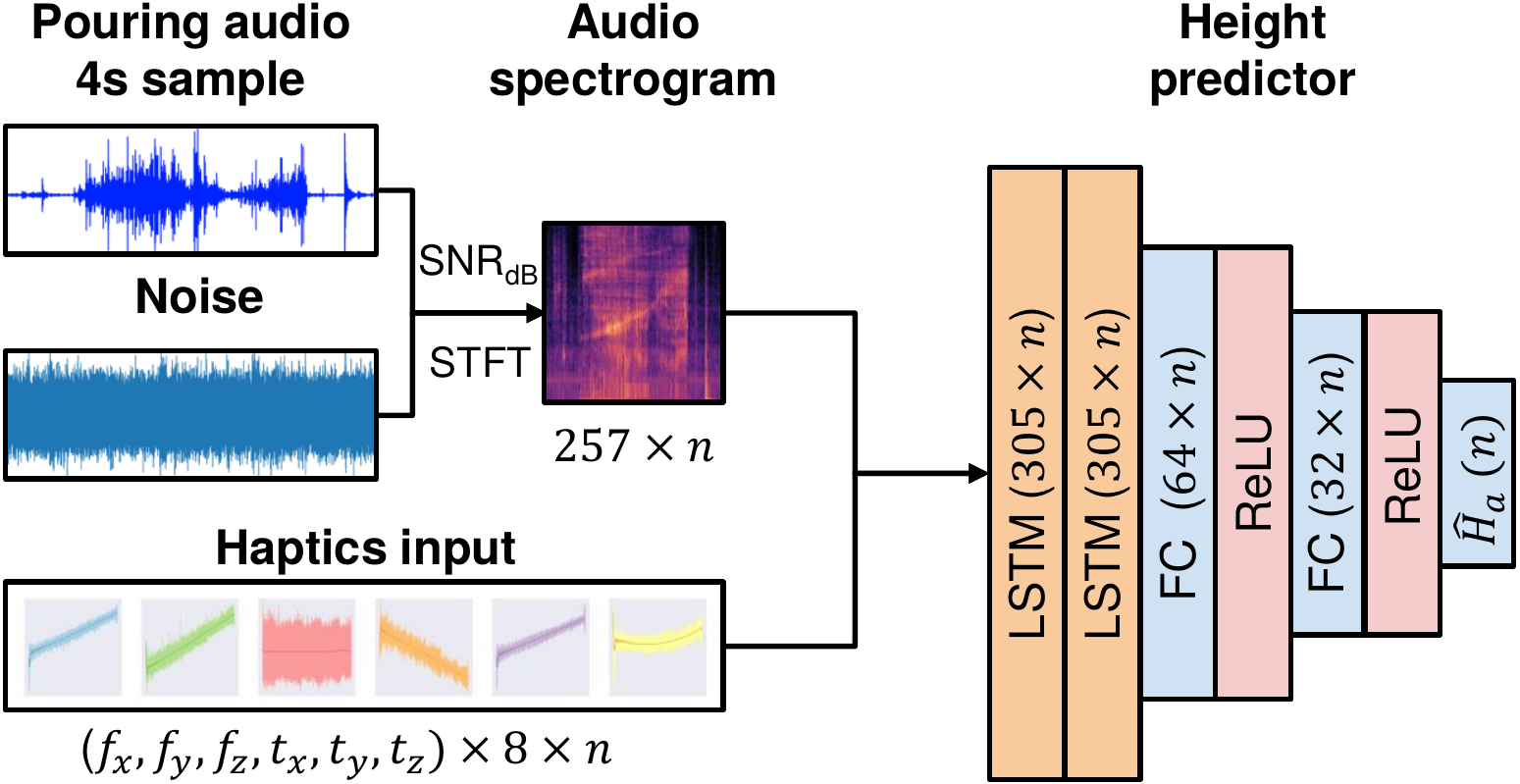}
    \caption{MP-Net architecture. The raw audio data is transformed into a spectrogram with 257 descriptors.
Then the encoder module (a recurrent neural unit) is progressively fed each time slice of the audio-frequency spectrogram and the corresponding haptic data.
Finally, the height predictor module produces the 1D length of the air column of the target containers.
    The blue rectangle denotes a fully-connected layer followed by a batch normalization layer and a rectified linear unit. $n = 251$ for a 4\,s pouring sample.
}
    \vskip -0.15in
    \label{fig:net}
\end{figure}

The pouring process is a typical sequential problem and the predicted length has a temporal relationship. Intuitively, we choose a recurrent network~\cite{cho2014properties} as our model architecture.
%using long short-term memory (LSTM)~\cite{cho2014properties} units
The multimodal network architecture MP-Net is shown in Fig.~\ref{fig:net}.
In detail, we utilize the long short-term memory (LSTM) unit to process the multimodal inputs.
We mix audio clips with noise using Equation~(\ref{eq3})
and transform it into a $257 \times n$ matrix using Short-Time Fourier Transform (STFT).
The haptic data is processed to a $6 \times 8 \times n$ matrix
($6$ F/T sensor channels, and $8$ F/T samples arrive during each $16$\,msec STFT interval)
and then concatenated with the audio data to a $305\times n$ matrix for input.
Here, $n$ is the number of the time slice, and we get $n = 251$ when using audio clips
of $4$ seconds duration from our dataset.
Then each time slice of the fusion data is progressively fed into the encoder module (2-layer LSTM unit) to a layer of 56 recurrent features $A_h$.

Besides, it is crucial to find a well-suited ground truth in supervised learning.
Due to the fact that when the length of the air column gets shorter in an organ pipe, the air vibrates faster and the resonance frequency of the air increases~\cite{french1983vino, webster2010use}, it is more indicative to choose the length of the air column $H_a$ as the ground truth of our model instead of the liquid height.
Thus, the height predictor (a 2-layer multilayer perceptron) in MP-Net takes the recurrent vector $A_h$ as input and performs a regression of the length of the air column $H_a$.
The height predictor is supervised with a mean squared error (MSE) loss $\mathcal{L}_{height}$

\begin{equation}
\label{heightloss}
\mathcal{L}_{height} = \|\hat{H}_a - H_a\|^{2}\text{.}
\end{equation}

In addition, an auxiliary $\mathcal{L}_{mono}$ is introduced to enforce the decrease of the estimated length of the air column over time $t$
\begin{equation}
\label{monoloss}
\mathcal{L}_{mono} = \sum\limits_t[\max(0, (\hat{H}_{a_{t+1}} - \hat{H}_{a_t}))]\text{.}
\end{equation}

\noindent
\textbf{Overall loss.} Combining with $\mathcal{L}_{height}$ and $\mathcal{L}_{mono}$, the complete training objective for MP-Net is defined by $\mathcal{L}_{mp}$
\begin{equation}
\label{audioloss}
\mathcal{L}_{mp}(\theta)=\mathcal{L}_{height} + \alpha \cdot \mathcal{L}_{mono}\text{,}
\end{equation}
where $\alpha$ is a hyperparameter for balancing these two loss functions.
In our implementation, we set it to $0.01$ for the best performance via some preliminary experiments.

\section{EXPERIMENT}
\subsection{MP-Net Evaluation}
\label{network}

We examined our proposed multimodal network MP-Net against the following baseline methods:
1) MP-Net*: the same network architecture as MP-Net but trained without adding noise to the audio data, 
2) AP-Net: only audio branch of MP-Net and training on augmented audio data with noise, 
3) AP-Net*: same network architecture as AP-Net without audio augmentation~\cite{liang2019AudioPouring},
4) FT-Net: only force/torque branch of MP-Net that only takes force/torque data as input.
All the models mentioned above were trained on our whole dataset which contains pouring samples collected from three different target containers.

\begin{figure}[t]
    \centering
    % \subfigure[]
    {\includegraphics[height=0.23\textwidth]{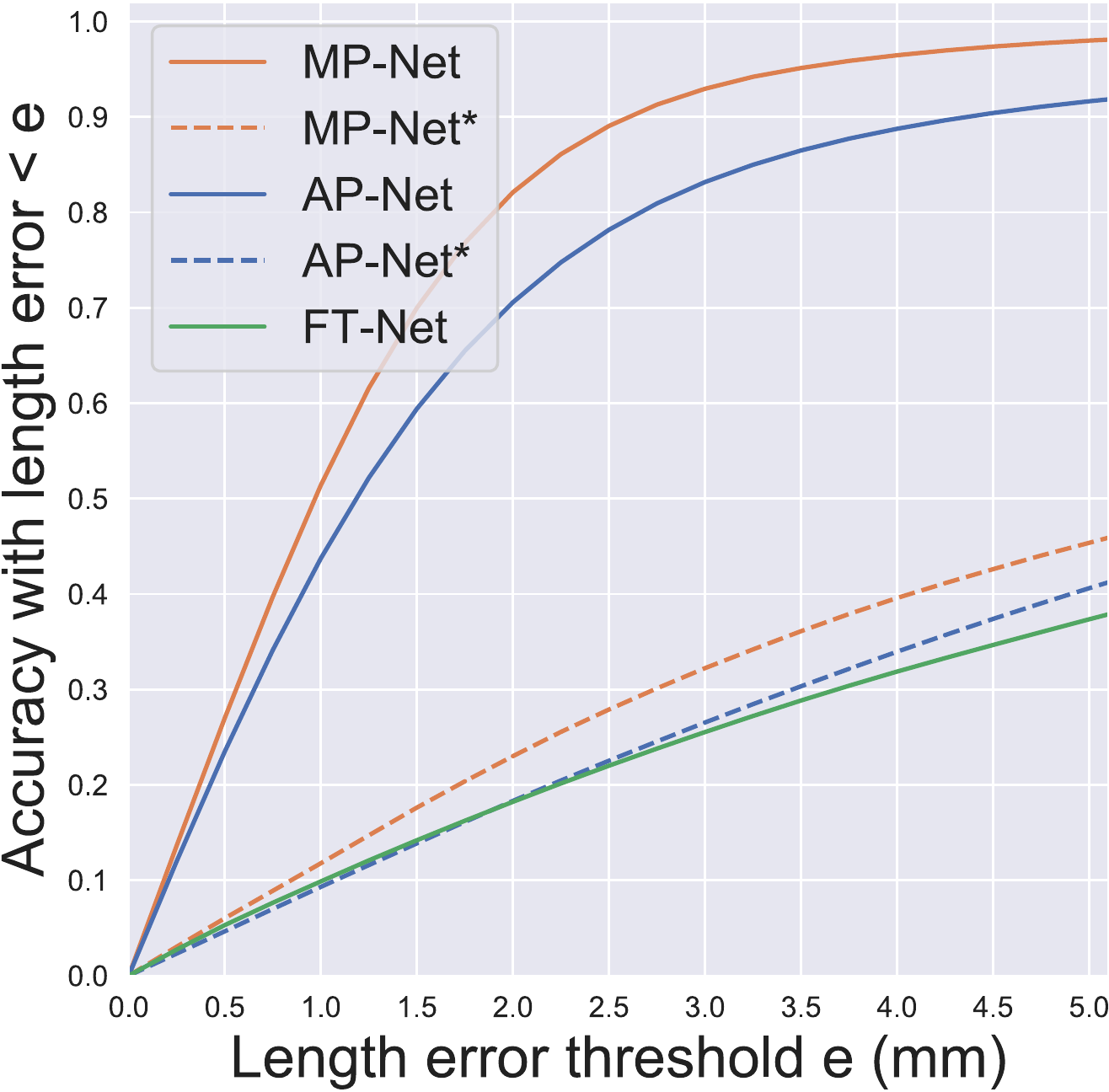}
    % \label{fig:network_threshold}
    }
    \hskip 2mm
    % \subfigure[]
    {\includegraphics[height=0.23\textwidth]{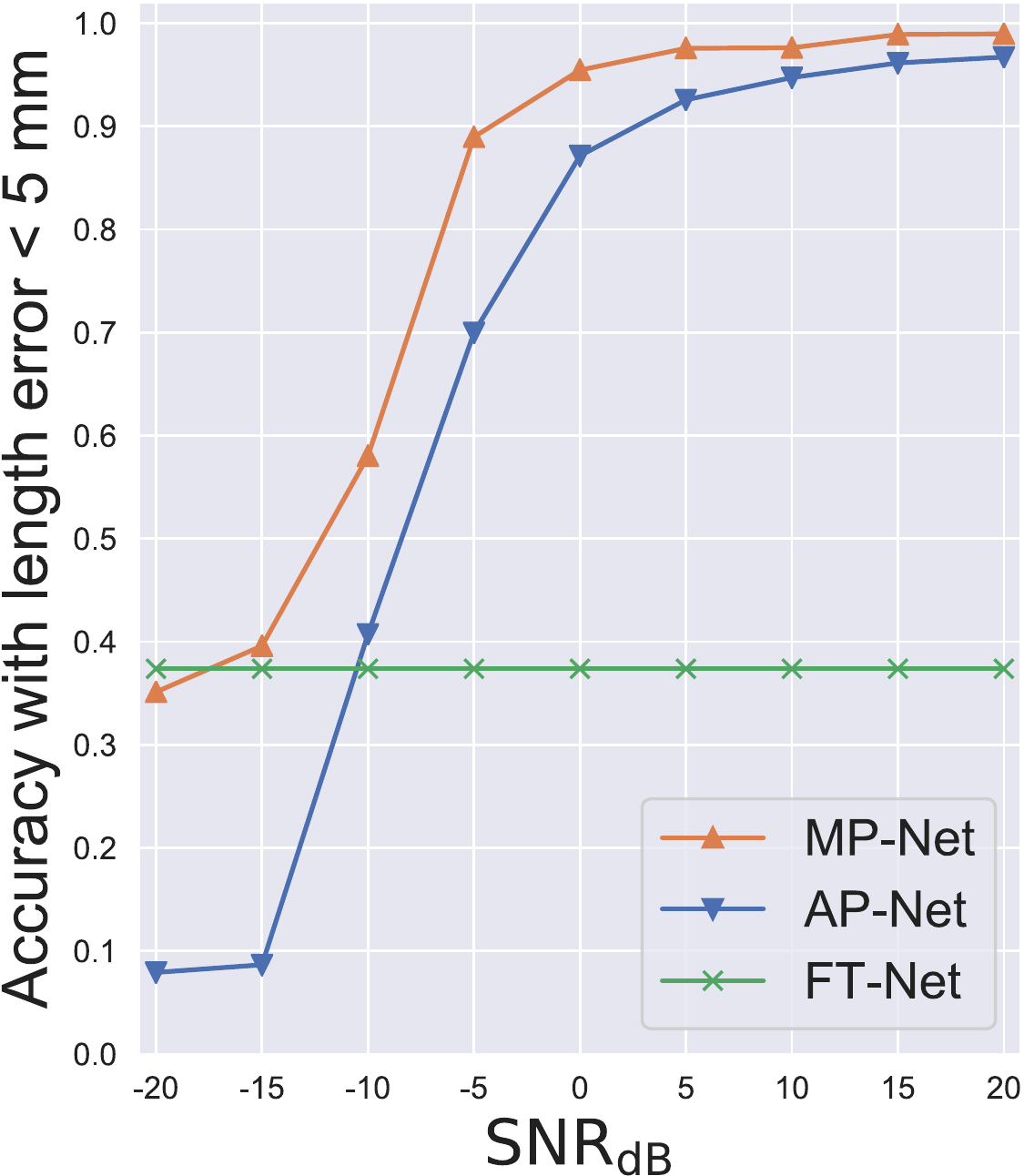}
    % \label{fig:network_snr}
    }
    \vspace{-2mm}
    \caption{Network evaluation results of MP-Net and four baselines. 
    % \subref{fig:network_threshold}
    (a, left) The fraction of the input sequence whose length prediction error is below a threshold between five models. 
    (b, right) The fraction of the input sequence whose length prediction error is below 5\,mm between MP-Net, AP-Net, FT-Net.
    MP-Net and AP-Net trained and tested on nine datasets, where the only difference between each dataset is the noise level of audio data. }
    \label{fig:network_eval}
    \vskip -2mm
\end{figure}

First, we evaluated MP-Net and four baselines using the fraction of the input sequences 
whose length prediction error $|\hat{H}_a - H_a| $ is below a threshold $e$.
We trained MP-Net and AP-Net on noisy audio, with noise levels from the $\rm SNR_{dB}$ set [0, 5, 10, 15, 20].
Except for FT-Net, MP-Net and the other three baseline models were then tested on audio data with $\rm SNR_{dB}=5$ noise level.
Fig.~\ref{fig:network_eval}(a) shows that MP-Net has a stable advantage of about 8\% over AP-Net 
over the full range of tested length error thresholds, 
which indicates that the force data provide a positive gain to the training result.
The performances of MP-Net* and AP-Net* are both poor compared to MP-Net and AP-Net, respectively, which indicates that augmenting audio data also contributes a lot to the networks.
FT-Net was not trained on audio and has learned to predict the air column length averaged over all containers.

To compare the robustness of MP-Net, AP-Net, FT-Net with regard to audio noise,
we evaluate the percentage of input sequences whose length prediction error stays below 5\,mm,
when both the training and the testing data were augmented with the same noise levels.
Each model was trained and tested on audio data augmented with 9 different $\rm SNR_{dB}$ levels from the set [-20, -15, -10, -5, 0, 5, 10, 15, 20] respectively.
Fig.~\ref{fig:network_eval}(b) illustrates that the accuracy of MP-Net and AP-Net increases with $\rm SNR_{dB}$.
The accuracy curve of FT-Net is straight because force/torque data is independent of audio noise.
We can see that MP-Net has at least 27\% higher accuracy than AP-Net when $\rm SNR_{dB} \le -15$, and has 8\% higher accuracy when $\rm SNR_{dB} = 0$.
Note that the performance of AP-Net degrades quickly for  $\rm SNR_{dB} \le -15$, 
which indicates that the noise interferes with the audio and the network cannot gain useful information from it anymore.
Nevertheless, with the help of force/torque data, MP-Net can still perform comparably to FT-Net in such a noisy situation.
This phenomenon proves that MP-Net has learned to predict liquid height by conditioning on both audio and haptic input.

\subsection{Robotic Experiments}
To verify and compare the reliability and robustness of the proposed MP-Net in robotic pouring tasks, we carried out eight evaluation experiments on
different target containers,
different pouring heights,
different source containers,
different noise levels,
varying positions of the noise source,
different initial liquid heights,
different types of liquid,
and different noise sources.

All experiments used both MP-Net and AP-Net to compare their real-world performance, except for the different source containers experiments.
To adapt to different noisy environments, 
MP-Net and AP-Net were trained on noisy audio data with noise levels $\rm SNR_{dB}$ from 0-20.
The experiment setup was the same as in our dataset collection setup 
and the initial water level of the source container was random.
MP-Net took the latest 4\,s sequence of audio and haptic data.
While the sequence length of the input signal was less than 4\,s, all the input signals were fed into the network.
Once the estimated length of the air column was smaller than the desired one, 
the robot immediately 
stopped pouring.
To compensate for the liquid poured out during the stopping motion, a corresponding correction was applied.

\subsubsection{Evaluation of Different Target Containers}
We used nine different target containers for robot experiments.
The first five rows in Table~\ref{tab:containers} illustrate the properties of these nine containers, where
containers 1-3 are included in the training dataset while 4-9 are only used for testing.
We kept the distance between the target containers and the microphone the same as in our original dataset.
Then we set the desired length of the air column to 60\,mm.
We synthesized a $\rm SNR_{dB}\!=\!5$ audio signal by playing~the~PR2 robot noise from a pair of loud\-speakers lo\-cated at positions 1\&1 as shown in Fig.~\ref{fig:pos}.
Here, 1\&1 means that both loudspeakers were placed at position 1.
We repeated the experiments five times for each container.

\renewcommand{\arraystretch}{.95}
\begin{table*}[ht]
    % \centering
    \begin{center}
        \caption{The properties of nine target containers and controlling results of pouring water into a desired length of the air column.}
        \vskip -0.15in
        \begin{tabular}{m{1.5cm}<{\centering}*{9}{m{1.3cm}<{\centering}}}
            \multicolumn{10}{c}{}\\
            \toprule
            %\hlineB{2}
            ID & 1 & 2 & 3 & 4 & 5 & 6 & 7 & 8 & 9 \\
            \midrule
            %\hlineB{2}
            \raisebox{-0.15cm}{Container} & \raisebox{-0.4cm}{\includegraphics[height=0.0582\textwidth]{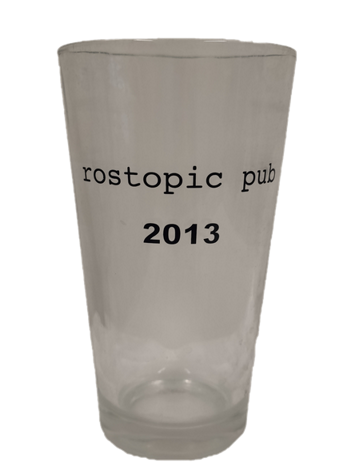}}& \raisebox{-0.4cm}{\includegraphics[height=0.0582\textwidth]{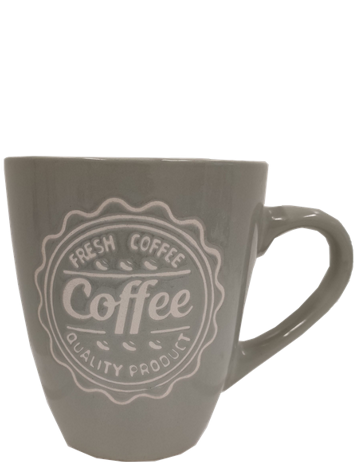}} & \raisebox{-0.4cm}{\includegraphics[height=0.0582\textwidth]{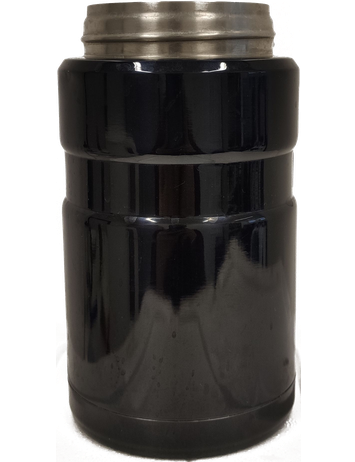}} & \raisebox{-0.4cm}{\includegraphics[height=0.0582\textwidth]{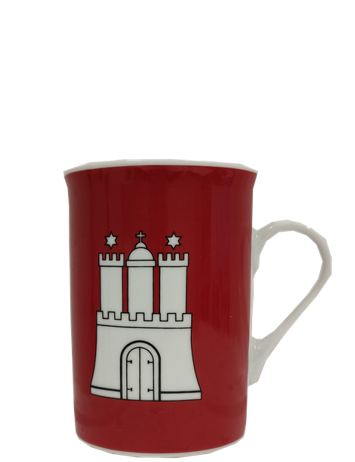}} & \raisebox{-0.4cm}{\includegraphics[height=0.0582\textwidth]{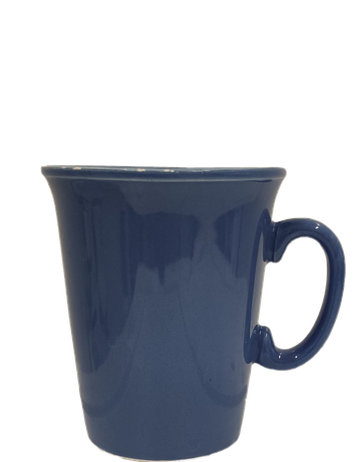}} & \raisebox{-0.4cm}{\includegraphics[height=0.0582\textwidth]{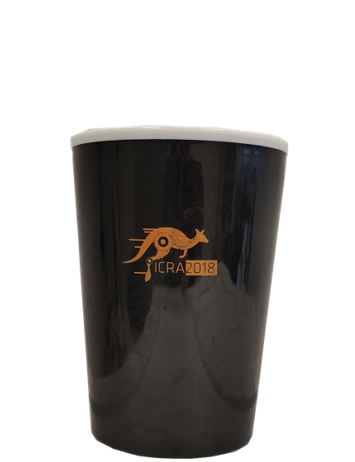}} & \raisebox{-0.4cm}{\includegraphics[height=0.0582\textwidth]{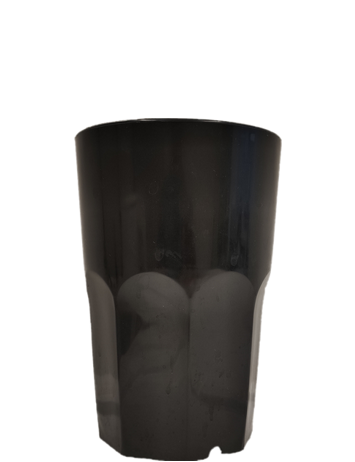}} & \raisebox{-0.4cm}{\includegraphics[height=0.0582\textwidth]{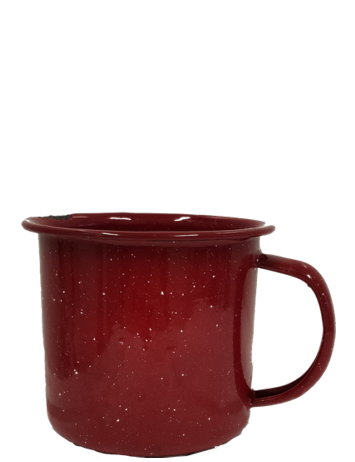}}& \raisebox{-0.4cm}{\includegraphics[height=0.0582\textwidth]{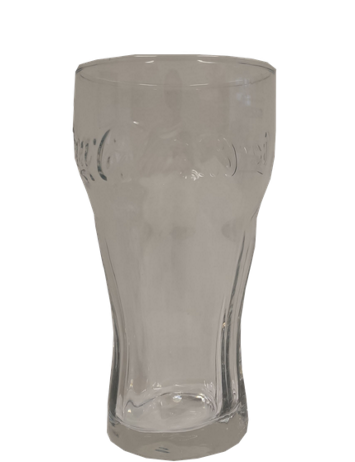}}\\
            \midrule
            %\hline
            Material & Glass & Ceramics & Steel & Ceramics & Ceramics & Plastic & Plastic & Enamel & Glass\\
            \midrule
            %\hline
            Height (mm) &127 &99 &150 &97 &94 &103 &115 &78 &135 \\
            \midrule
            %\hline
            Volume (ml) &460 &428 &766 &310 &298 &416 &408 &382 &358 \\
            \bottomrule
            %\hlineB{2}
            \multicolumn{10}{c}{Error in volume (ml) with $\rm SNR_{dB} = 5 $, $H_a=60$\,mm } \\
            \toprule
            %\hlineB{2}
            MP-Net &$15.8 \pm 14.0$ &$11.5 \pm 8.1$ &$6.3 \pm 6.1$ &$16.6 \pm 19.0$ &$8.2 \pm 9.3$&$12.8 \pm 11.5$ &$23.7 \pm 18.7$ &$8.5 \pm 9.6$ &$13.9 \pm 4.4$ \\
            \midrule
            %\hline
            AP-Net &$23.1 \pm 8.2$ &$7.5 \pm 8.2$ &$6.0 \pm 4.7$ &$25.3 \pm 14.5$ &$5.2 \pm 2.3$ &$14.3 \pm 10$ &$46.6 \pm 70.6$ &$41.6 \pm 10.4$ &$26.6 \pm 45.4$ \\
            \bottomrule
            %\hlineB{2}
            % \vskip 0.3in
        \end{tabular}
        \label{tab:containers}
    \end{center}
    \vskip -0.2in
\end{table*}

\begin{figure}[t]
    \centering
    \subfigure[]
    {\includegraphics[height=0.21\textwidth]{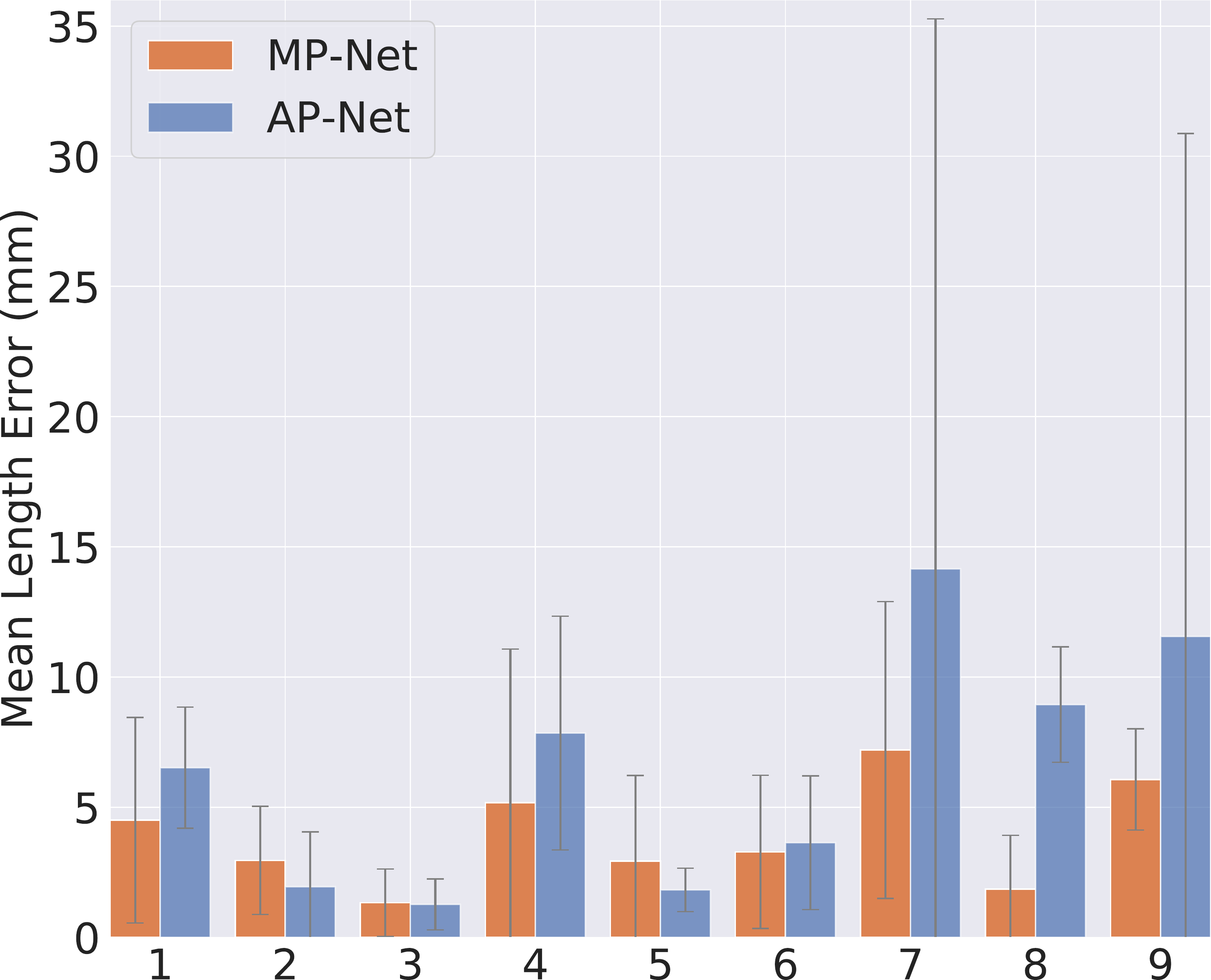}
    \label{fig:diff_cups}
    }
    \subfigure[]
    {\includegraphics[height=0.21\textwidth]{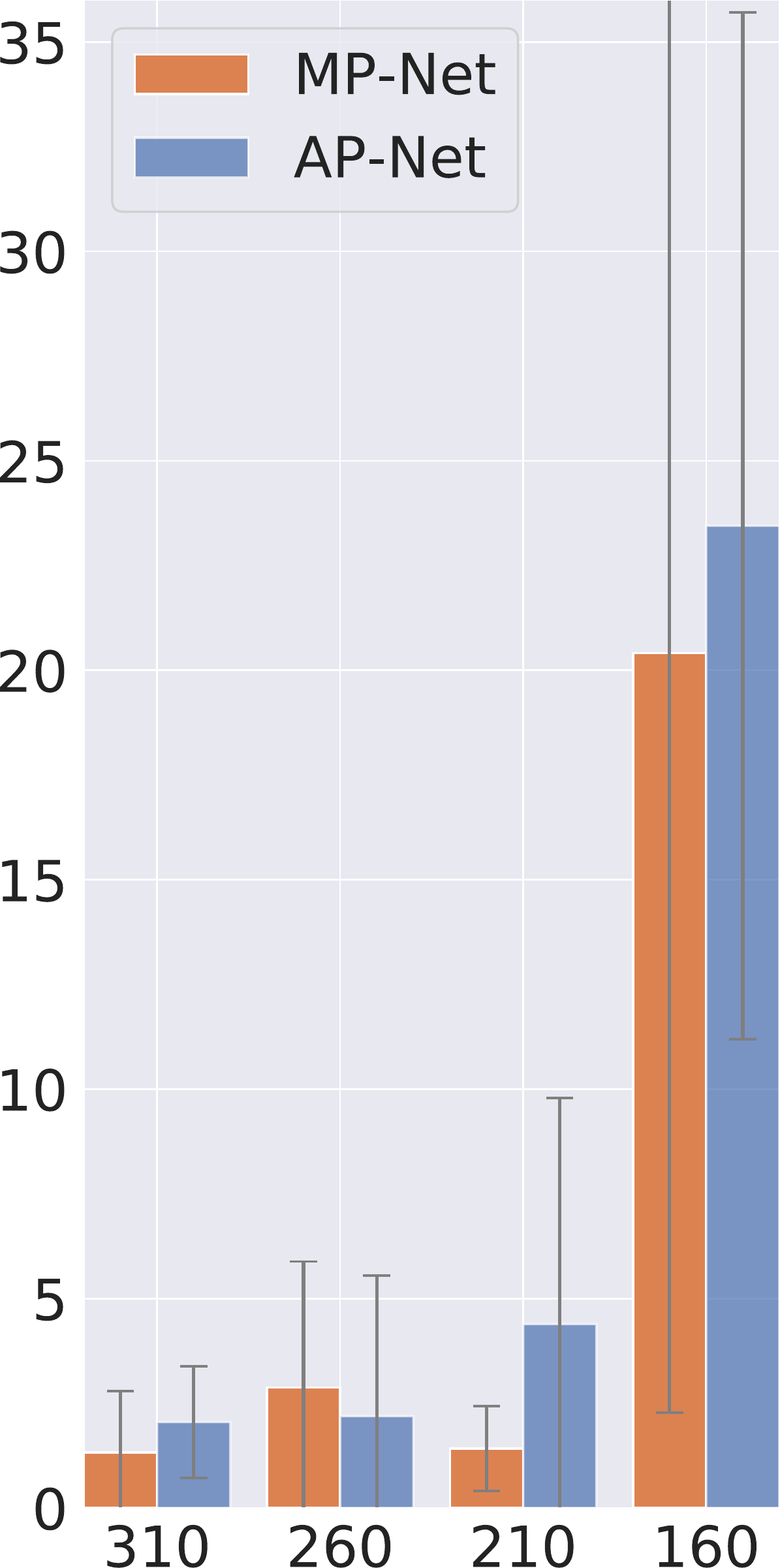}
    \label{fig:diff_pouring_height}
    }
    \subfigure[]
    {\includegraphics[height=0.21\textwidth]{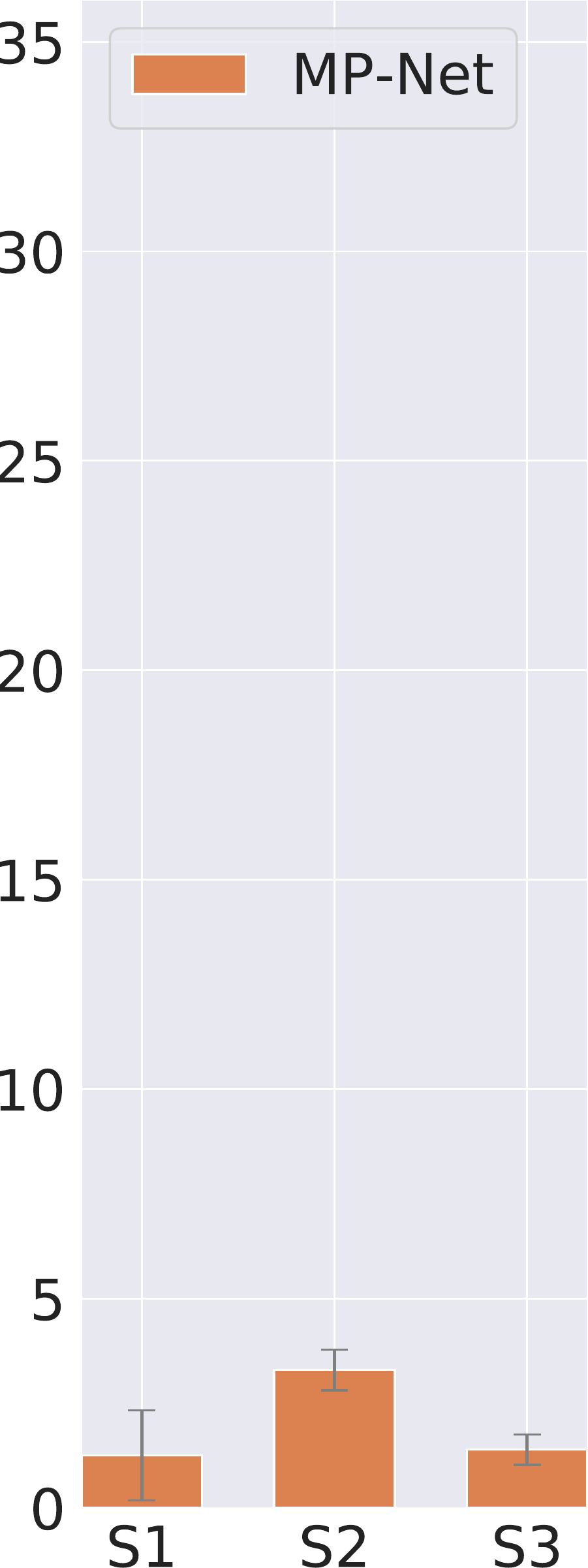}
    \label{fig:diff_source_container}}
    \vskip -2mm
    \caption{ In the following robot experiments, the $\rm SNR_{dB}$ were set to 5\,dB. Robot experiment result \subref{fig:diff_cups} on pouring water into different target containers ($H_a=60$\,mm, ID 1-3 are in the training dataset, ID 4-9 are novel),   \subref{fig:diff_pouring_height} on four different pouring heights (pouring height 310\,mm was the height for dataset collection, $H_a=40$\,mm),
    \subref{fig:diff_source_container} on one known source container S1 and two novel source containers S2, S3 ($H_a=40$\,mm).
    }
    \vskip -0.25in
\end{figure}

Fig.~\ref{fig:diff_cups} displays that the absolute mean errors of the liquid height are below 8\,mm and the standard deviations are below 4\,mm for both MP-Net and AP-Net among the known target containers.
Container 3 performs best for both networks due to the stainless steel material which makes the crispest sound.
Furthermore, we converted the height error of each cup to a weight error, shown in rows seven and eight of Table~\ref{tab:containers}.
We can see that AP-Net performs well on known containers and even outperforms MP-Net a little on containers 2 and 3.
Regarding the unseen containers, especially for container 7 and 9, MP-Net exhibits a stronger generalization ability than AP-Net in a noisy environment.

\subsubsection{Evaluation of Different Pouring Heights}
In this experiment, we placed the source container at four different heights from the set [310, 260, 210, 160]\,mm respectively. 
The height of the source container is the vertical distance from the mouth center of the source container to the scale plane. 
We carried out five robot experiments on each height using target container 2 when $\rm SNR_{dB} = 5$.
Fig. ~\ref{fig:diff_pouring_height} indicates that our algorithm performs well for the three higher heights but not at the lowest height. 
As the pouring height decreases, the volume of the pouring sound also decreases. With the source container at height 160\,mm 
the pouring sound is too low, and haptics only is not sufficient for the network to perceive the height of the air column.

\subsubsection{Evaluation of Different Source Containers}
The source container influences the flow-rate and the initial force/torque data.
Therefore, we tested two novel source containers S2 (44\,g) and S3 (484\,g), which are shown in the left side of Fig.~\ref{fig:dataset_setup}, to compare with the container S1 (64\,g) used for network training and all other experiments. We kept the pouring height at 310\,mm, and the other experimental setup was the same as in the evaluation of the different pouring height.
Fig.~\ref{fig:diff_source_container} suggests that different source containers hardly affect network performance.

\subsubsection{Evaluation of Varying Noise Conditions}
To further verify the performance of our MP-Net model in different noise conditions, we implemented a set of experiments using target container 2 under six different noise 
levels of \mbox{[-5, 0, 5, 10, 15, 20]} $\rm SNR_{dB}$. 
The loudspeakers were again located at positions 1\&1.
    For each $\rm SNR_{dB}$ level, we tested five different target lengths of air column, namely [40, 50, 60, 70, 80]\,mm.
    We carried out five robot experiments on each audio and target length condition.

\begin{figure}[t]
    \includegraphics[width=0.48\textwidth]{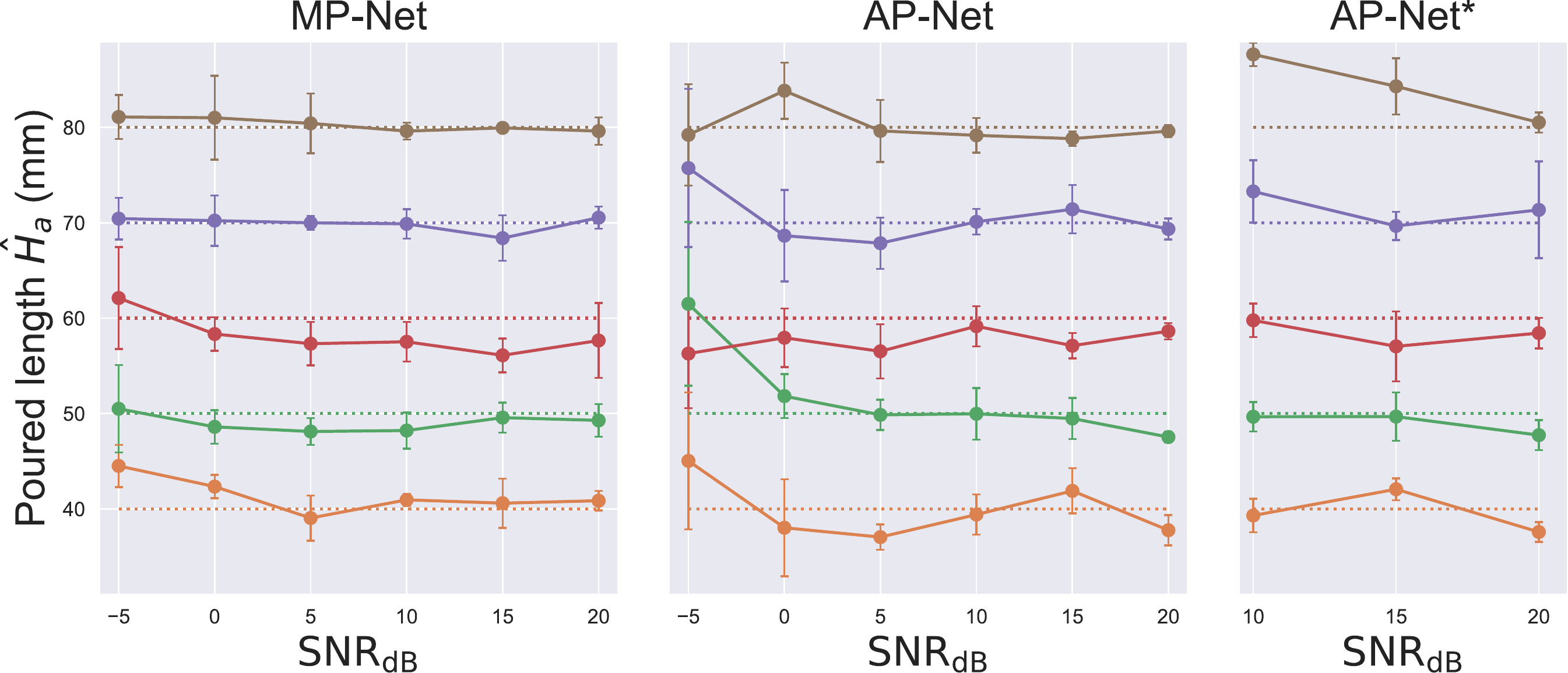}
    \caption{Robot experiment results of the performance of MP-Net, AP-Net, and AP-Net* in environments with different levels of noise (measured by $\rm SNR_{dB}$). We evaluate with five different target liquid heights and the results are demonstrated in five different colors, respectively. The dashed lines represent the desired lengths of the air column, while the solid dots (with error bars) show the actual ones when the pouring terminates.}
    \label{fig:three_model}
    \vskip -0.25in
\end{figure}

    As visualized in Fig.~\ref{fig:three_model}, MP-Net has a substantial advantage over AP-Net when $\rm SNR_{dB} = -5$, which further indicates the advantages of multimodal fusion.
In this experiment, we also tested AP-Net* under different noise conditions.
AP-Net* performed well while $\rm SNR_{dB}\ge10$, but when $\rm SNR_{dB}<10$, the robot either stopped pouring immediately or overfilled the target containers.
Therefore we did not list the experiment results of AP-Net* tested on audio with $\rm SNR_{dB}<10$.

\begin{figure*}[t]
    \centering
    \subfigure[]
    {\includegraphics[height=0.24\textwidth]{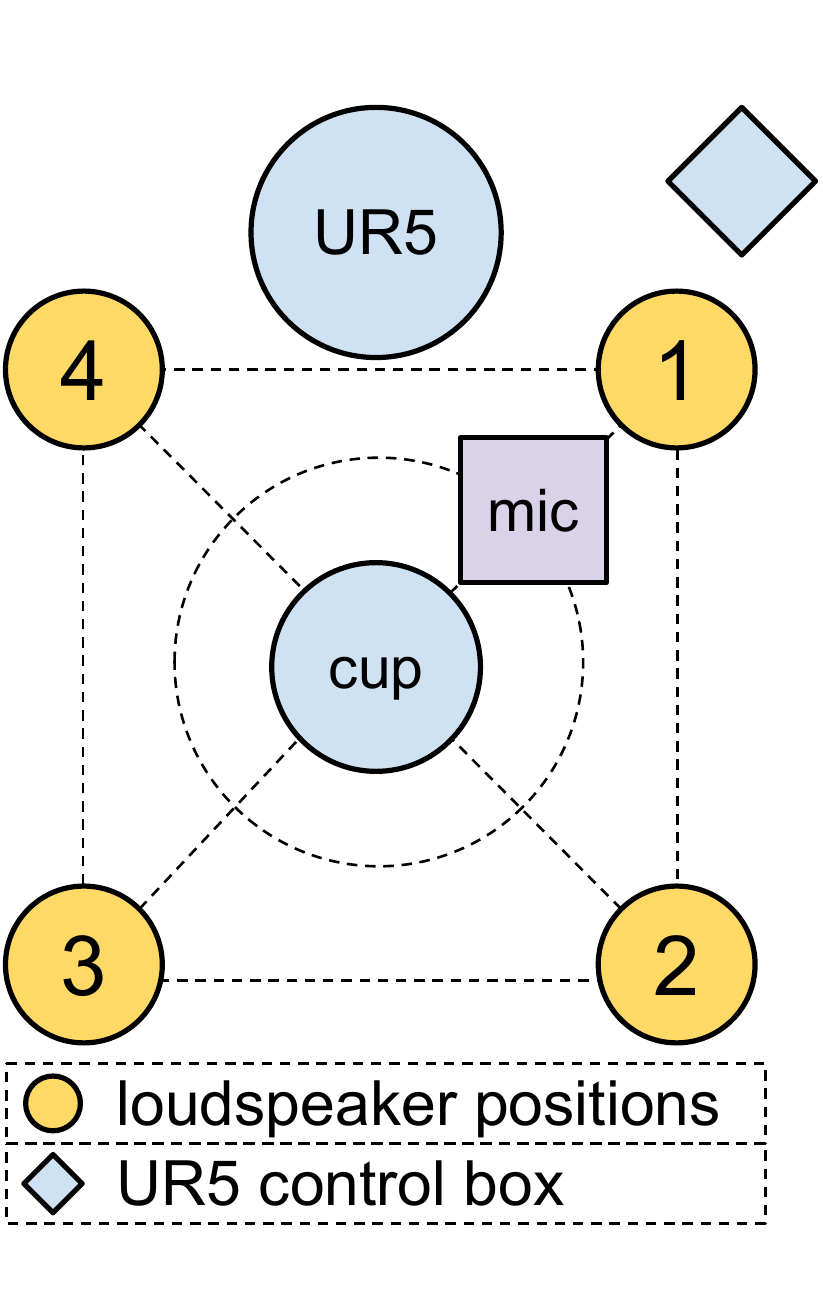}
    \label{fig:pos}\hspace{.37cm}}
    \subfigure[]
    {\includegraphics[height=0.23\textwidth]{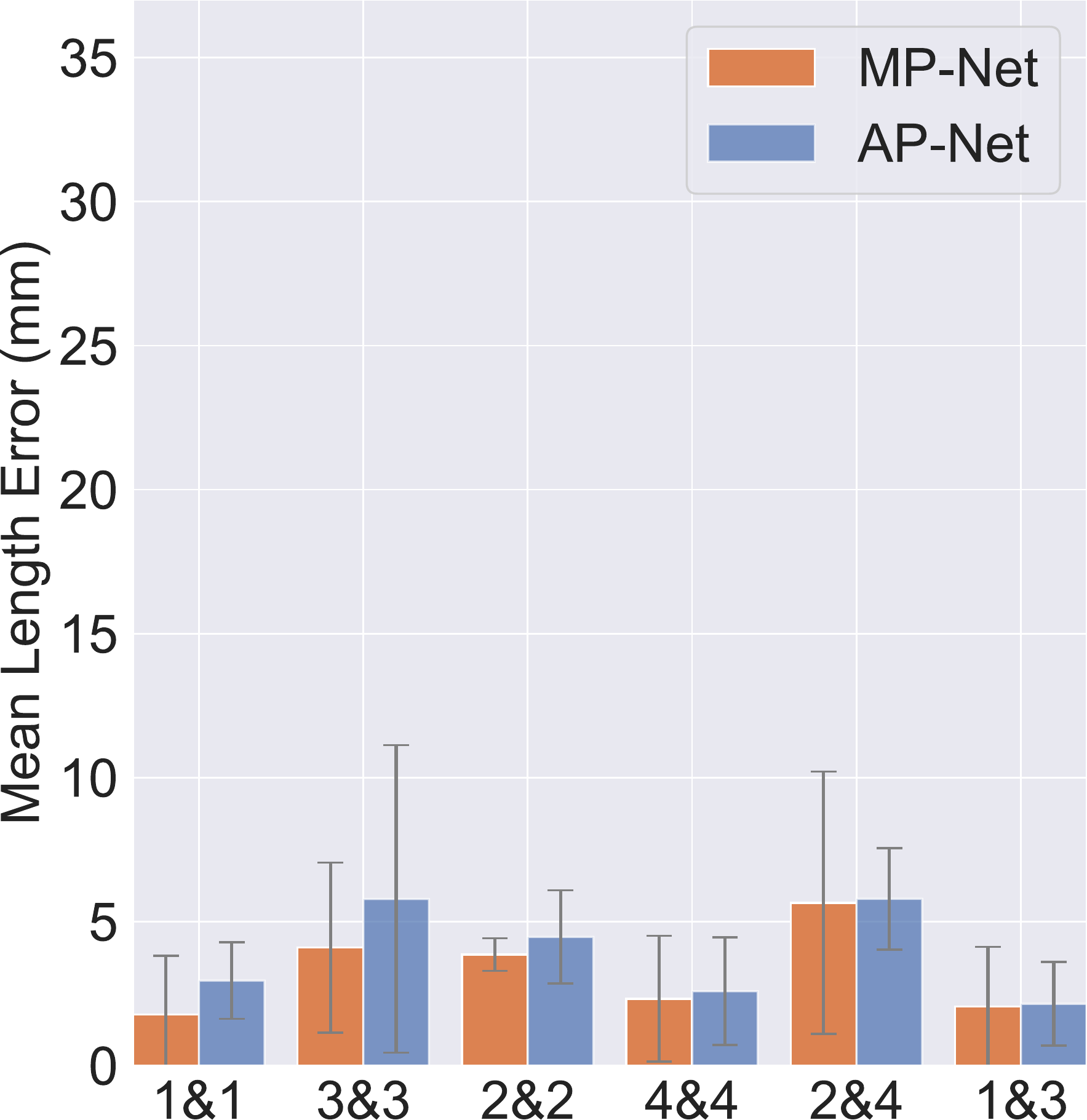}
    \label{fig:pos_eval}\hspace{.48cm}}
    \subfigure[]
    {\includegraphics[height=0.23\textwidth]{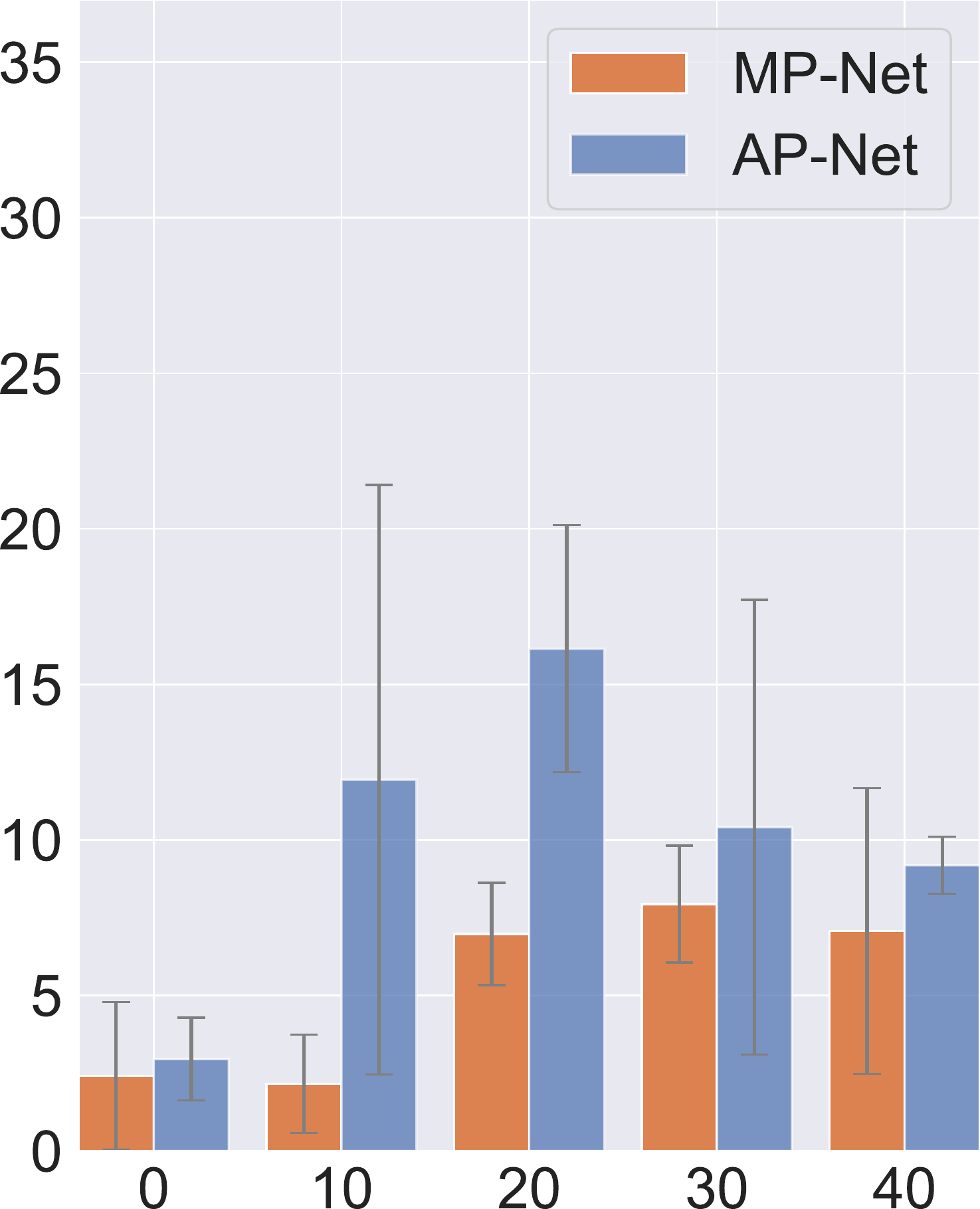}
    \label{fig:inital}\hspace{.48cm}}
    \subfigure[]
    {\includegraphics[height=0.23\textwidth]{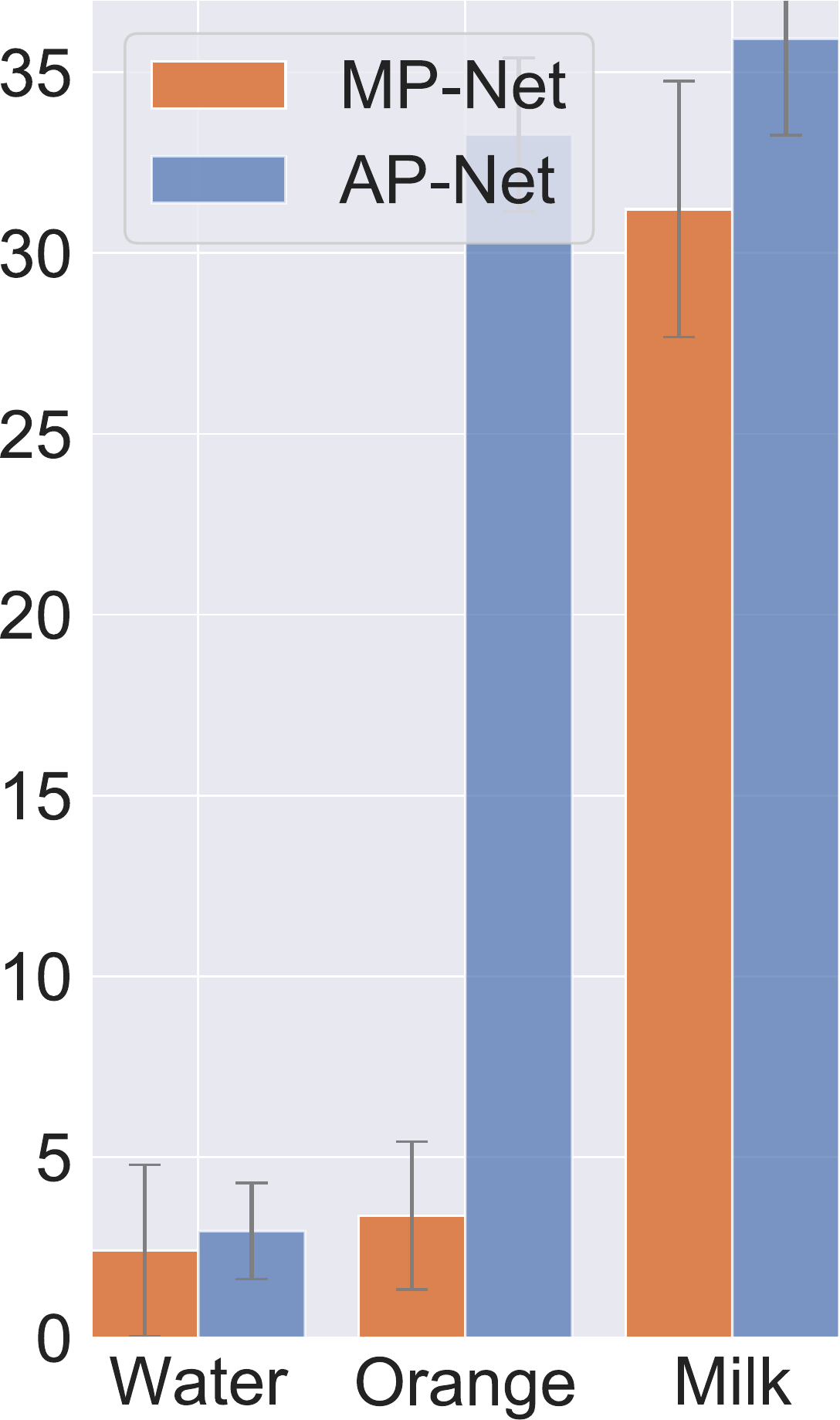}
    \label{fig:diff_liquid}\hspace{.48cm}}
    \subfigure[]
    {\includegraphics[height=0.23\textwidth]{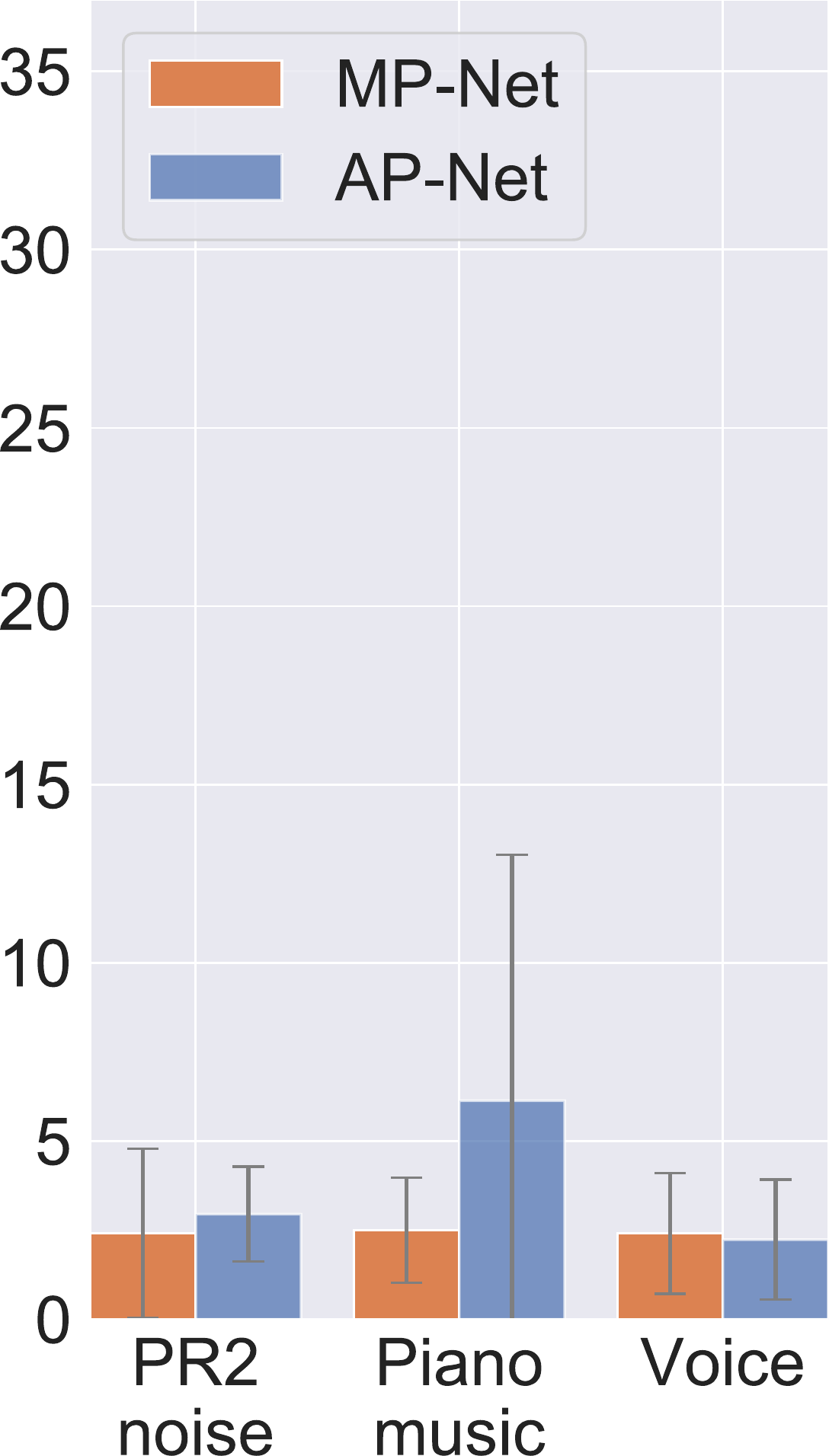}
    \label{fig:diff_noise_type}}
    \vskip -1mm
    \caption{
    \subref{fig:pos} Schematic diagram of 4 different loudspeaker positions relative to the target containers, the UR5 robot position, and the control box of the UR5 robot.
     Evaluation results of \subref{fig:pos_eval} six combinations of two loudspeakers positions, \subref{fig:inital} varying initial heights of the source container, \subref{fig:diff_liquid} different types of liquids and \subref{fig:diff_noise_type} different types of noise sources. In these experiments, the $\rm SNR_{dB}$ was set to 5\,dB. The target height $H_a$ was set to 40\,mm.
    }
    \vskip -0.2in
\end{figure*}

    \subsubsection{Evaluation of Varying Positions of Noise Source}
To assess whether MP-Net is sensitive to the direction of the noise source, we set up six different position combinations [1\&1, 2\&2, 3\&3, 4\&4, 2\&4, 1\&3] of the two loudspeakers.

The two loudspeakers played a synthetic $\rm SNR_{dB}\!= \!5$ noise signal at each position. 
We used the target container 2 and a desired air column length of 40\,mm.
The other experimental setup was the same as in the evaluations of the different target containers.
Then at each combined position of the two loudspeakers, we poured water five times.
As shown in Fig.~\ref{fig:pos_eval}, 
when the loudspeakers are at position 1\&1, both models perform best as the loudspeakers are behind the microphone. 
MP-Net generalizes better than AP-Net to the different positions of the loudspeakers due to the lower consistent mean height error among all tested positions.

\subsubsection{Evaluation of Varying Initial Liquid Height in Target Containers}
In this experiment, we poured water into the target container 2, starting from five different initial liquid heights [0, 10, 20, 30, 40]\,mm.
We tested five times from each initial level.
We put two loudspeakers at 1\&1 positions and kept the other test setups the same as in the evaluations of the varying direction of noise sources.
The results in Fig.~\ref{fig:inital} demonstrate that MP-Net is again more robust than AP-Net.
Force and torque data yield a meaningful indication of how much water was poured out.

\subsubsection{Evaluation of Different Types of Liquid}
    We conducted pouring experiments with different liquids: pure water, orange juice and 1.8\% fat milk.
    We used the same experimental setting as in the evaluations of different microphone positions and poured each type of liquid for five times. 
    As manifested in Fig.~\ref{fig:diff_liquid}, MP-Net can generalize to common household liquids like water and orange juice while AP-Net cannot handle the task of 
    pouring orange juice well under $\rm SNR_{dB} = 5$.
    However, similar to \cite{liang2019AudioPouring}, due to the high viscosity of milk, both models cannot generate correct height prediction.

\subsubsection{Evaluation of Different Types of Noise Sources}
    We also assessed our model with three noise types: PR2 robot noise, human voices and a continuous piece of piano music.
    The human voice is represented by discrete sounds of a man counting numbers in English.
    We poured water under each type of noise five times.
    All experimental settings were the same as in the evaluations of different types of liquid.
    Fig.~\ref{fig:diff_noise_type} shows that MP-Net is not affected by different noise types, but the accuracy of AP-Net has a small fluctuation under a musical disturbance.

\subsection{Shape Prediction of Target Containers}
\begin{figure*}[ht]
    
    \begin{minipage}[b]{0.15\textwidth} 
    \centering
    \includegraphics[height=3.5cm]{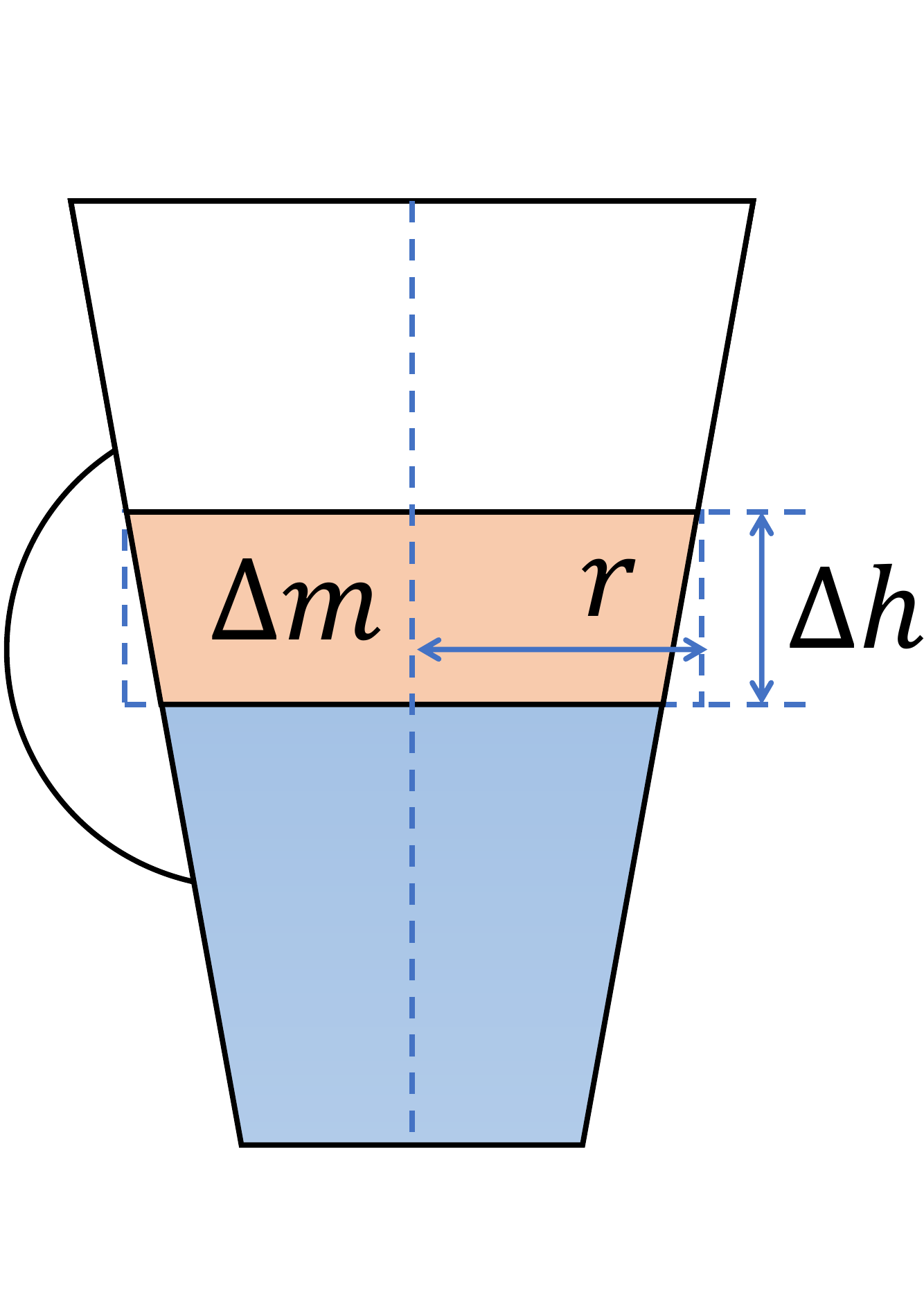}
    \end{minipage}
    \begin{minipage}[b]{0.84\textwidth}
    \centering
    \includegraphics[height=3.5cm]{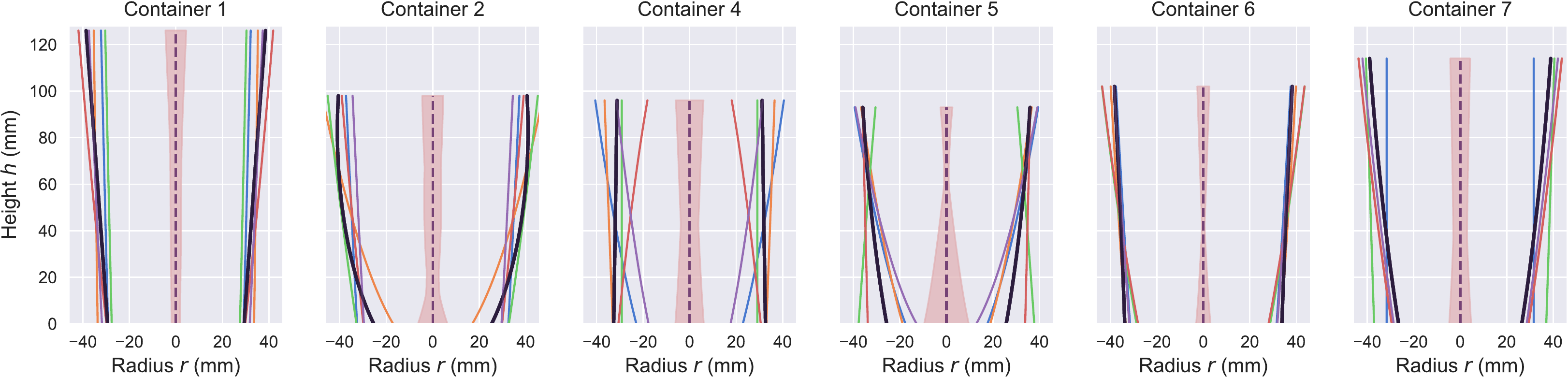}

    \end{minipage}
    \caption{(a, left) Schematic diagram of a symmetric container.
    In a specified time interval $\Delta t$, the change in mass $\Delta m$ can be determined by the F/T sensor and the change in height $\Delta h$ can be derived 
    through MP-Net. Then the radius $r$ at each height can be calculated to form an edge-profile of this container.
    % \label{fig:edge_profile}
    (b, right) Prediction result of estimating the target container shape. The black curve is the ground truth, while the five different colored curves are the estimated target container shape in five trials.
    The mean error of the estimated container radius at different heights is plotted in the middle of each subplot (the shaded magenta area).
    % \label{fig:predict_bottle_shape}
    \label{bottle_shape}
    }
    \vskip -0.15in
\end{figure*}

In this section, we applied MP-Net to predict the shape of symmetric target containers.
In this case, the edge profile is sufficient to describe the shape of the containers, 
which is determined by the correlation between height and radius~\cite{kennedy2019autonomous}.
Fig.~\ref{bottle_shape}(a) shows a volume profile filled with liquid of density $\rho$, where $\Delta V, \Delta m, \Delta h$ are the poured liquid volume, and the weight and liquid height differences during a time interval $\Delta t$ respectively.
Assuming that $\Delta t$ is very small, then $\Delta m$ can be calculated by approximating the shape of $\Delta V$ as a cylinder,
\vskip -0.2in

\begin{equation}
    \label{eq7}
    \Delta m = \rho \Delta V 
    = \rho \pi r^2 \Delta h
\end{equation}
We can determine $\Delta m$ by the $(f_x, f_y, f_z)$ values from the force/torque sensor and $\Delta h$ through our neural network output $\hat{H}_a$.
In the robot experiments, the frequency of $\Delta m$ and $\hat{H}_a$ was 500\,Hz and 12\,Hz, respectively. To get a smooth and accurate estimation of the container shape, we used a quadratic function to fit the scatter points,
\begin{equation}
    \label{eq8}
    r(h) \approx \sqrt{\frac{\Delta m}{\Delta h}\frac{1}{\rho\pi}}
\end{equation}
For target containers 1, 2, 4, 5, 6, 7, we conducted five trials of the experiment in which the robot pours into these target containers.
We recorded the realtime estimation of $\hat{H}_a$ and force data into a rosbag. 
When the target container was filled to about 90\% of its total height, we stopped the pouring and the recording.
Using the data from these rosbags, we calculated the edge-profiles of the target containers.
In Fig.~\ref{bottle_shape}(b), 
the thick black curves are the ground truth profiles, while five colored curves around black curves depict the experimental results.
The magenta area in the middle of each target container visualizes the mean error of the radius prediction.
As expected, the mean radius estimation error is highest for an empty container,
when our recursive network cannot yet rely on its memory but stabilizes as the liquid level rises.
Due to the restriction to quadratic functions, 
the reconstruction works best for containers with low edge curvature (such as containers 1, 7).

\section{Conclusion and Future Work}  % (0.25 page)
In this paper, we motivate the need for combining audio and haptic information for robot pouring tasks.
We recorded a robot pouring dataset that includes 300 complete robot pouring sequences with audio and force/torque data. 
We propose a novel audio-haptic recurrent deep network (MP-Net) trained on this dataset that predicts liquid height in realtime.
The multimodal perception system is systematically tested across four baselines and a wide range of robotic pouring experiments in a noisy environment.
The results substantiate that MP-Net is quite robust against noise and against changes in different tasks and varying environments.
Finally, the multimodal nature of our network lets us reconstruct the shape of the target container.
The dataset and associated software are public and are available at
 \href{https://lianghongzhuo.github.io/MultimodalPouring}{https://lianghongzhuo.github.io/MultimodalPouring}.

One surprising limitation of our approach is the poor generalization to liquids like milk or fruit juices, which would be considered quite similar to water by many humans, while the pouring noises are actually quite different.
Training on different liquid types would improve network performance, but MP-Net will still fail in situations where the auditory signal is too weak.
Another issue is our use of raw force/torque data as the network input,
which changes significantly for different grasp types and pouring motions. This could be resolved by training on many grasps, or simply by feeding preprocessed weight data into the network.

For future work, using audio and haptic information for dynamic control of robotic pouring would be an exciting research direction.

\bibliographystyle{IEEEtran} % (0.5 page, about 20 refs)
\bibliography{IEEEabrv,ref}

\end{document}